\pdfoutput=1

\documentclass[11pt]{article}

\usepackage[]{acl}

\usepackage{times}
\usepackage{latexsym}

\usepackage[T1]{fontenc}

\usepackage[utf8]{inputenc}

\usepackage{microtype}

\usepackage{microtype}
\usepackage{booktabs}
\usepackage{multirow}
\usepackage{graphicx}
\usepackage{caption}
\usepackage[normalem]{ulem}
\usepackage[utf8]{inputenc}
\usepackage{dingbat}
\usepackage{wrapfig}
\usepackage{amssymb}
\usepackage{amsfonts}
\usepackage{amsmath}
\usepackage{amssymb}
\usepackage{amsthm}
\usepackage{mathrsfs}
\usepackage{adjustbox}
\usepackage{bbold}
\usepackage[font=small]{caption}
\usepackage[linesnumbered,ruled]{algorithm2e}
\usepackage{url}
\usepackage{subcaption}
\usepackage{xspace}

\title{Expanding Pretrained Models to Thousands More Languages\\via Lexicon-based Adaptation}

\author{
\textbf{Xinyi Wang}$^{1}$ \quad
\textbf{Sebastian Ruder}$^{2}$ \quad
\textbf{Graham Neubig}$^{1}$
\\
$^{1}$Language Technology Institute, Carnegie Mellon University \\
$^{2}$Google Research \\
\texttt{xinyiw1@cs.cmu.edu,ruder@google.com,gneubig@cs.cmu.edu}
}

\begin{document}
\maketitle

\begin{abstract}
The performance 
of multilingual pretrained models is highly dependent on the availability of monolingual or parallel text present in a target language. Thus, the majority of the world's languages cannot benefit from recent progress in NLP as they have no or limited textual data.
To expand possibilities of using NLP technology in these under-represented languages, 
we systematically study strategies that relax the reliance on conventional language resources through the use of bilingual lexicons, an alternative resource with much better language coverage.
We analyze different strategies to synthesize textual or labeled data using lexicons, and how this data can be combined with monolingual or parallel text when available.
For 19 under-represented languages across 3 tasks, our methods lead to consistent improvements of up to 5 and 15 points with and without extra monolingual  text respectively. Overall, our study highlights how NLP methods can be adapted to thousands more languages that are under-served by current technology.%
\footnote{Code and data are available at: \url{https://github.com/cindyxinyiwang/expand-via-lexicon-based-adaptation}.}
\end{abstract}
\section{Introduction}

Multilingual pretrained models~\citep{bert,xlm,xlmr} have become an essential method for cross-lingual transfer on a variety of NLP tasks~\citep{pires-etal-2019-multilingual,wu-dredze-2019-beto}. These models can be finetuned on annotated data of a down-stream task in a high-resource language, often English, and then the resulting model is applied to other languages. This paradigm is supposed to benefit under-represented languages that do not have annotated data. However, recent studies have found that the cross-lingual transfer performance of a language is highly contingent on the availability of monolingual data in the language during pretraining~\cite{xtreme}. Languages with more monolingual data tend to have better performance while languages not present during pretraining significantly lag behind.

\begin{figure}
    \centering
    \includegraphics[width=\columnwidth]{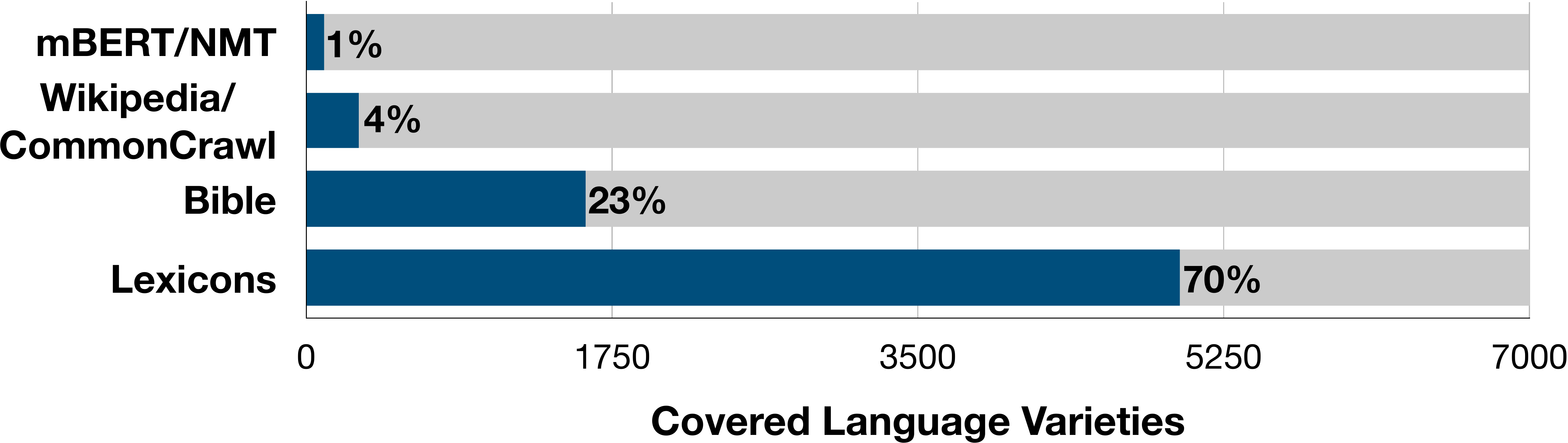}
    \caption{The percentage of the world's $\approx$7,000 languages covered by mBERT, monolingual data sources and lexicons.}
    \label{fig:languages}
\vspace{-2em}
\end{figure}
Several works propose methods to adapt the pretrained multilingual models to low-resource languages, but these generally involve continued training using monolingual text from these languages~\citep{wang-etal-2020-extending,chau-etal-2020-parsing,mad-x,unks2021}. Therefore, the performance of these methods is still constrained by the amount of monolingual or parallel text available, making it difficult for languages with little or no textual data to benefit from the progress in pretrained models. \citet{joshi-etal-2020-state} indeed argue that unsupervised pretraining makes the `resource-poor poorer'. 

\autoref{fig:languages} plots the language coverage of multilingual BERT \cite[mBERT;][] {bert}, a widely used pre-trained model, and several commonly used textual data sources.\footnote{Statistics taken from \citet{adapt-1600-languages} and \url{panlex.org}.} Among the 7,000 languages
in the world, mBERT only covers about $1\%$ of the languages while Wikipedia and CommonCrawl, the two most common resources used for pretraining and adaptation, only contain textual data from $4\%$ of the languages (often in quite small quantities, partially because language IDs are difficult to obtain for low-resource languages~\citep{language-id-wild}). \citet{adapt-1600-languages} show that continued pretraining of multilingual models on a small amount of Bible data can significantly improve the performance of uncovered languages. Although the Bible has much better language coverage of $23\%$, its relatively small data size and constrained domain limits its utility~(see \autoref{sec:g1_exp})---and $70\%$ of the world's languages do not even have this resource.
The failure of technology to adapt to these situations raises grave concerns regarding the fairness of allocation of any benefit that may be conferred by NLP to speakers of these languages \citep{joshi-etal-2020-state,blasi2021systematic}.
On the other hand, linguists have been studying and documenting under-represented languages for years in a variety of formats~\citep{gippert2006essentials}. Among these, bilingual lexicons or word lists are usually one of the first products of language documentation, and thus have much better coverage of the worlds' languages than easily accessible monolingual text, as shown in \autoref{fig:languages}.
There are also ongoing efforts to create these word lists for even more languages through methodologies such as ``rapid word collection''~\citep{rapidwordbrenda}, which can create an extensive lexicon for a new language in a number of days.
As \citet{bird-2020-decolonising} notes:
\begin{quote}
     After centuries of colonisation, missionary endeavours, and linguistic fieldwork, all languages have been identified and classified. There is always a wordlist. 
     $\ldots$
     In short, we do not need to “discover” the language ex nihilo (L1 acquisition) but to leverage the available resources (L2 acquisition).
\end{quote}

However, there are few efforts on understanding the best strategy to utilize this valuable resource for adapting pretrained language models. Bilingual lexicons have been used to synthesize bilingual data for learning cross-lingual word embeddings \cite{gouws-sogaard-2015-simple,ruder2019survey} and task data for NER via word-to-word translation~\citep{mayhew-etal-2017-cheap}, but both approaches precede the adoption of pre-trained multilingual LMs. 
\citet{khemchandani-etal-2021-exploiting} use lexicons to synthesize monolingual data for adapting LMs, but their experimentation is limited to several Indian languages and no attempt was made to synthesize downstream task data while \citet{Hu2021amber} argue that bilingual lexicons may hurt performance.

\begin{figure}
    \centering
    \includegraphics[width=0.8\columnwidth]{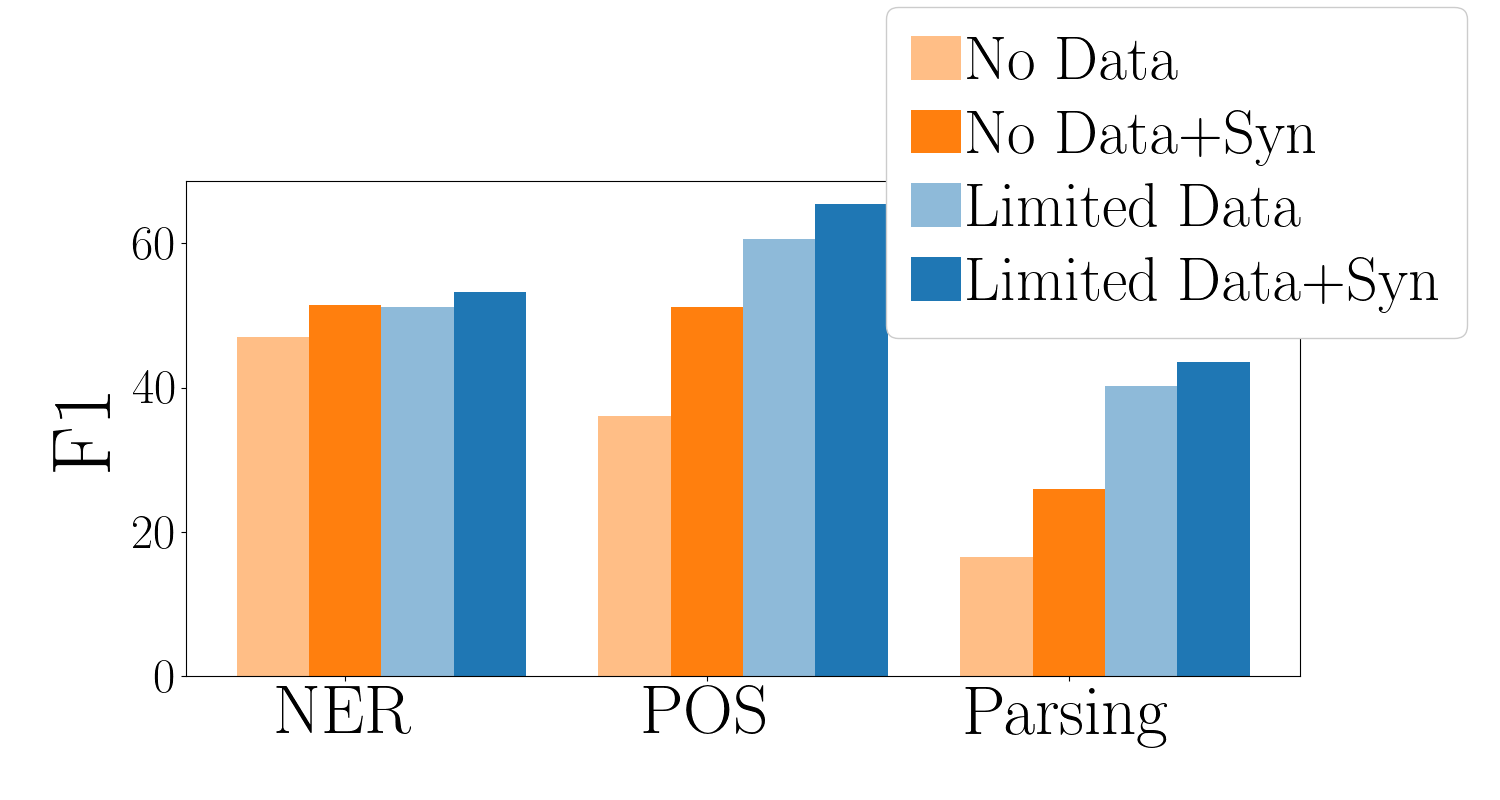}
    \vspace{-5mm}
    \caption{Results for baselines and adaptation using synthetic data for both resource settings across three NLP tasks.}
    \label{fig:improvement}
    \vspace{-5mm}
\end{figure}

In this paper, we conduct a systematic study of strategies to leverage this relatively under-studied resource of bilingual lexicons to adapt pretrained multilingual models to languages with little or no monolingual data.
Utilizing lexicons from an open-source database, we create synthetic data for both continued pretraining and downstream task fine-tuning via word-to-word translation.
Empirical results on 19 under-represented languages on 3 different tasks demonstrate that using synthetic data leads to significant improvements on all tasks~(\autoref{fig:improvement}), and that the best strategy depends on the availability of monolingual data~(\autoref{sec:g0_exp}, \autoref{sec:g1_exp}). We further investigate methods that improve the quality of the synthetic data through a small amount of parallel data or by model distillation.

\section{Background}

We focus on the cross-lingual transfer setting where the goal is to maximize performance on a downstream task in a target language $T$. Due to the frequent unavailability of labeled data in the target language, a pretrained multilingual model $M$ is typically fine-tuned on labeled data in the downstream task $\mathcal{D}_{label}^S = \{ ( x_i^S, y_i^S ) \}_{i=1}^N $ in a source language $S$ where $x_i^S$ is a textual input, $y_i^S$ is the label, and $N$ is the number of labeled examples. The fine-tuned model is then directly applied to task data $\mathcal{D}_{test}^T = \{ x_i^T, y_i^T \}_i $  in language $T$ at test time.\footnote{We additionally examine the few-shot setting where some task data $\mathcal{D}_{label}^T$ in $T$ is available for fine-tuning in \autoref{sec:analysis}.} The performance on the target language $T$ can often be improved by further adaptation of the pretrained model. 

\subsection{Adaptation with Text}
There are two widely adopted paradigms for adapting pretrained models to a target language using monolingual or parallel text.
\paragraph{MLM} Continued pretraining on monolingual text $ \mathcal{D}_{mono}^T = \{ x_i^T \}_i$ in the target language  ~\citep{Howard2018,dont-stop-pretrain-gururangan-etal-2020} using a masked language model~(MLM) objective has proven effective for adapting models to the target language \citep{mad-x}. Notably, \citet{adapt-1600-languages} show that using as little as several thousand sentences can significantly improve the model's performance on target languages not covered during pretraining.

\paragraph{Trans-Train} 
For target languages with sufficient parallel text with the source language $ \mathcal{D}_{par}^{ST} = \{ (x_i^S, x_i^T ) \}_i $, one can train a machine translation (MT) system that translates data from the source language into the target language. Using such an MT system, we can translate the labeled data in the source language $\mathcal{D}_{label}^S$ into target language data $\widehat{\mathcal{D}}_{label}^T = \{ ( \widehat{x}_i^T, y_i^S ) \}_{i=1}^N $, and fine-tune the pretrained multilingual model on both the source and translated labeled data $\mathcal{D}_{label}^S \cup \widehat{\mathcal{D}}_{label}^T$. This method often brings significant gains to the target language, especially for languages with high-quality MT systems~\citep{xtreme,xtreme-r}.

\subsection{Challenges with Low-resource Languages}
Both methods above require $\mathcal{D}_{mono}^T$ or $\mathcal{D}_{par}^{ST}$ in target language $T$, so they cannot be directly extended to languages without this variety of data.
\citet{joshi-etal-2020-state} classified the around 7,000 languages of the world into six groups based on the availability of data in each language. The two groups posing the biggest challenges for NLP are:
\begin{description}
    \item[``The Left-Behinds,''] languages with virtually no unlabeled data. We refer to this as the \emph{No-Text} setting.
    \item[``The Scraping-Bys,''] languages with a small amount of monolingual data. We refer to this as the \emph{Few-Text} setting.
\end{description}
These languages make up $85\%$ of languages in the world, yet they do not benefit from the development of pretrained models and adaptation methods due to the lack of monolingual and parallel text. 
In this paper, we conduct a systematic study of strategies directly targeted at these languages.

\section{Adapting to Under-represented Languages Using Lexicons}
\begin{figure}
    \centering
    \includegraphics[width=\columnwidth]{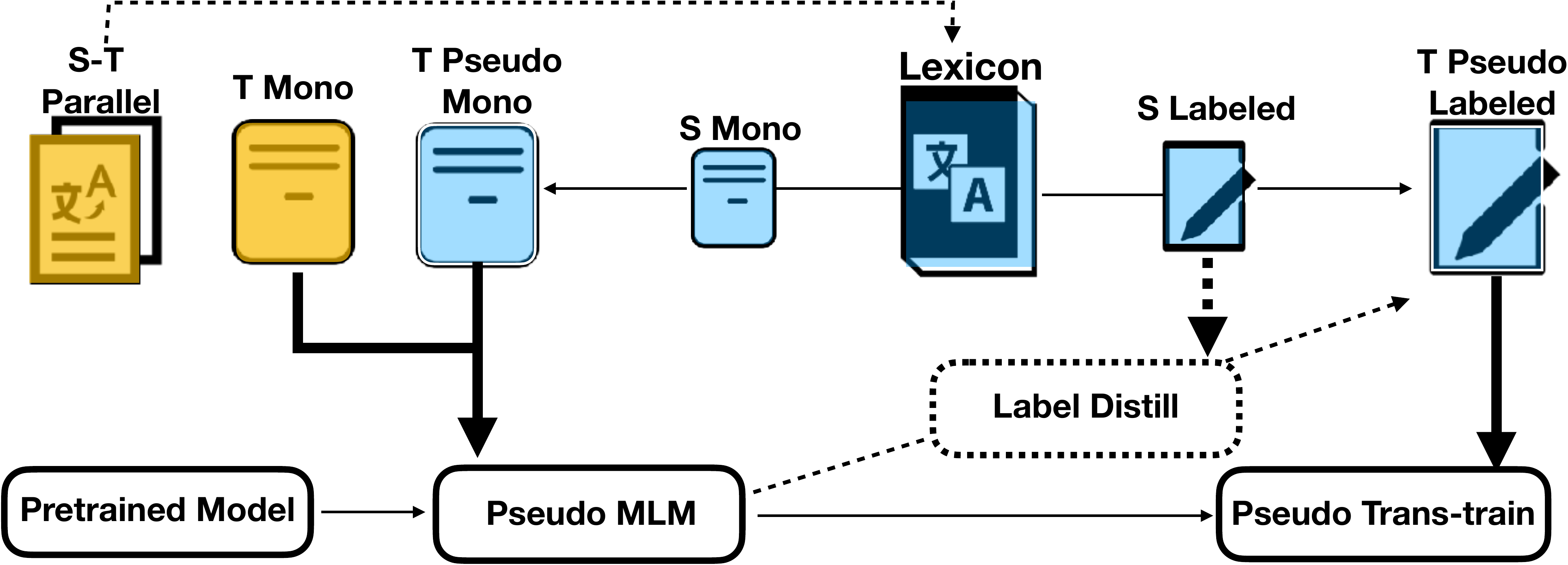}
    \caption{Pipelines for synthesizing data for \textcolor{cyan}{both No-text and Few-text} settings and utilizing extra data for the \textcolor{brown}{Few-Text} setting. Solid lines indicate adaptation methods and dashed lines are synthetic data refinement methods.}
    \label{fig:method}
    \vspace{-5mm}
\end{figure}

Since the main bottleneck of adapting to under-represented languages is the lack of text, we adopt a data augmentation framework~(illustrated in \autoref{fig:method}) that leverages bilingual lexicons, which are available for a much larger number of languages.

\subsection{Synthesizing Data Using Lexicons}

Given a bilingual lexicon $\mathcal{D}_\text{lex}^{ST}$ between the source language $S$ and a target language $T$, we create synthetic sentences $\widetilde{x}_i^T$ in $T$ using sentences $x_i^S$ in $S$ via word-to-word translation, and use this synthetic data in the following adaptation methods.

\paragraph{Pseudo MLM}
Using monolingual text $\mathcal{D}_{mono}^S = \{ x_i^S \}_i$, we generate pseudo monolingual text $\widetilde{\mathcal{D}}_{mono}^T = \{\widetilde{x}_i^T\}_i$ for $T$ by replacing the words in $x_i^S$ with their translation in $T$ based on the lexicon  $\mathcal{D}_\text{lex}^{ST}$. We keep the words that do not exist in the lexicon unchanged, so the pseudo text $\widetilde{x}_i^T$ can include words in both $S$ and $T$. We then adapt the pretrained multilingual model on $\widetilde{\mathcal{D}}_{mono}^T$ using the MLM objective.
For the Few-Text setting where some gold monolingual data $\mathcal{D}_{mono}^T$ is available, we can train the model jointly on the pseudo and the gold monolingual data $\widetilde{\mathcal{D}}_{mono}^T \cup \mathcal{D}_{mono}^T $.

\paragraph{Pseudo Trans-train} Given the source labeled data $\mathcal{D}_{label}^S = \{ ( x_i^S, y_i^S ) \}_{i=1}^N$, for each text example $x_i^S$ we use $\mathcal{D}_\text{lex}^{ST}$ to replace the words in $x^S_i$ with its corresponding translation in $T$, resulting in pseudo labeled data $\widetilde{\mathcal{D}}_{label}^T = \{ ( \widetilde{x}_i^T, y_i^S ) \}_{i=1}^N$. We keep the original word if it does not have an entry in the lexicon. We then fine-tune the model jointly on both pseudo and gold labeled data $ \widetilde{\mathcal{D}}_{label}^T \cup \mathcal{D}_{label}^S$.

Since these methods only require bilingual lexicons, we can apply them to both No-Text and Few-Text settings. We can use either of the two methods or the combination of both to adapt the model.  

\paragraph{Challenges with Pseudo Data}
Our synthetic data $\widetilde{\mathcal{D}}^T$ could be very different from the true data $\mathcal{D}^T$ because the lexicons do not cover all words in $S$ or $T$, and we do not consider morphological or word order differences between $T$ and $S$.\footnote{In fact, we considered more sophisticated methods using morphological analyzers and inflectors, but even models with relatively broad coverage~\citep{anastasopoulos-neubig-2019-pushing} did not cover many languages we used in experiments.} 
Nonetheless, we find that this approach yields significant improvements in practice~(see \autoref{tab:main_results}). We also outline two strategies that aim to improve the quality of the synthetic data in the next section. 

\begin{table*}[]
    \centering
    \resizebox{0.9\textwidth}{!}{
    \begin{tabular}{l|l}
    \toprule
       eng $x^S \in \mathcal{D}_{mono}^S$ & Anarchism calls for the abolition of the state , which it holds to be undesirable , unnecessary , and harmful .  \\
       Pseudo Mono $\widetilde{x}^T \in \widetilde{\mathcal{D}}_{mono}^T$  & Anarchism calls \textcolor{gray}{g$\hbar$al} \textcolor{gray}{il} abolition \textcolor{gray}{ta’ il stat , lima hi} holds \textcolor{gray}{g$\hbar$al tkun} undesirable , \textcolor{gray}{bla bzonn , u} harmful . \\
       \midrule
       eng $x^S \in \mathcal{D}_{label}^S$ & I suspect the streets of Baghdad \textcolor{red}{will} \textcolor{red}{look} as if a war is looming this week .  \\
       Pseudo Labeled $\widetilde{x}^T \in \widetilde{\mathcal{D}}_{label}^T$ & \textcolor{gray}{jien iddubita il} streets \textcolor{gray}{ta’} Bagdad \textcolor{red}{xewqa} \textcolor{red}{hares} \textcolor{gray}{kif jekk a gwerra} is looming \textcolor{gray}{dan ġimg$\hbar$a} . \\
       Pseudo Labeled   $y^S \in \widetilde{\mathcal{D}}_{label}^T$  & PRON VERB DET NOUN ADP PROPN \textcolor{red}{AUX} \textcolor{red}{VERB} SCONJ SCONJ DET NOUN AUX VERB DET NOUN PUNCT \\ 
       Label Distilled $\widetilde{y}^T\in \widetilde{\mathcal{D}}_{distill}^T$    & PRON VERB DET NOUN ADP PROPN \textcolor{red}{NOUN} \textcolor{red}{NOUN} SCONJ SCONJ DET NOUN AUX VERB DET NOUN PUNCT \\
    \bottomrule
    \end{tabular}}
    \caption{Examples of pseudo monolingual data and pseudo labeled data for POS tagging for Maltese~(mlt). Words in red have different labels between the source language and the label distilled data. This is because ``xewqa'' in Maltese is a noun meaning ``desire,will'', while the word ``will'' is not used as a noun in the original English sentence.}
    \label{tab:pseudo_data_example}
\end{table*}
\subsection{Refining the Synthetic Data}
\paragraph{Label Distillation} 
The pseudo labeled data $\widetilde{\mathcal{D}}_{label}^T  = \{ ( \widetilde{x}_i^T, y_i^S ) \}_{i=1}^N$ is noisy because the synthetic examples $\widetilde{x}_i^T$ could have a different label from the original label $y_i^S$~(see \autoref{tab:pseudo_data_example}). To alleviate this issue, we propose to automatically ``correct'' the labels of pseudo data using a teacher model. Specifically, we fine-tune the pretrained multilingual model as a teacher model using only $\mathcal{D}_{label}^S$. We use this model to generate the new pseudo labeled data $\widetilde{\mathcal{D}}_{distill}^T = \{ ( \widetilde{x}_i^T, \widetilde{y}_i^T ) \}_{i=1}^N $ by predicting labels $\widetilde{y}^T_i$ for the pseudo task examples $\widetilde{x}^T_i$. 
We then fine-tune the pretrained model on both the new pseudo labeled data and the source labeled data $\widetilde{\mathcal{D}}_{distill}^T \cup \mathcal{D}_{label}^S$.

\paragraph{Induced Lexicons with Parallel Data} For the Few-Text setting, we can leverage the available parallel data $\mathcal{D}_{par}^{ST}$ to further improve the quality of the augmented data. Specifically, we use unsupervised word alignment to extract additional word pairs $\widetilde{\mathcal{D}}_{lex}^{ST}$ from the parallel data, and use the combined lexicon  $\widetilde{\mathcal{D}}_{lex}^{ST} \cup \mathcal{D}_{lex}^{ST}$ to synthesize the pseudo data.

\section{General Experimental Setting}
In this section, we outline the tasks and data setting used by all experiments. We will then introduce the adaptation methods and results for the No-Text setting in \autoref{sec:g0_exp} and the Few-Text setting in \autoref{sec:g1_exp}. 
\subsection{Tasks, Languages and Model} 
We evaluate on the gold test sets of three different tasks with relatively good coverage of under-represented languages: named entity recognition~(NER), part-of-speech~(POS) tagging, and dependency parsing~(DEP). We use two NER datasets: WikiAnn NER~\citep{pan-etal-2017-cross,rahimi-etal-2019-massively} and MasakhaNER~\citep{adelani2021masakhaner}. We use the Universal Dependency 2.5~\citep{nivre2018universal} dataset for both the POS and DEP tasks. 

We use English as the source language for all experiments. For each dataset, we use the English training data and select the checkpoint with the best performance on the English development set. For MasakhaNER, which does not have English training data, we follow \citet{adelani2021masakhaner} and use the CoNLL-2003 English NER training data. We run each fine-tuning experiment with 3 random seeds and report the average performance. For NER and POS tagging, we follow the data processing and fine-tuning hyper-parameters in  \citet{xtreme}. We use the Udify~\citep{udify-kondratyuk-straka-2019-75} codebase and configuration for parsing. 

\begin{table}[]
    \centering
    \resizebox{\columnwidth}{!}{
    \begin{tabular}{l|l|l|l|l}
    \toprule
      Language  & iso & Family & Task & Lex Count \\
    \midrule
      Acehnese   & ace & Austronesian & NER & 0.5k \\
      Bashkir   & bak &  Turkic & NER & 3.4k \\
      Crimean Turkish & crh & Turkic & NER & 4.4k \\
      Hakka Chinese  & hak &  Sino-Tibetan & NER & 8.5k \\
      Igbo   & ibo & Niger-Congo & NER & 3.6k \\
      Ilokano   & ilo & Austronesian  & NER & 4.0k \\
      Kinyarwanda  & kin &  Niger-Congo & NER & 4.7k\\
      Eastern Mari  & mhr &  Uralic & NER & 21.7k \\
      Maltese  & mlt & Afro-Asiatic & All & 1.0k \\
      Maori   & mri &  Austronesian & NER & 13.8k \\
      Hausa   & hau &  Niger-Congo & NER & 5.6k \\
      Wolof   & wol &  Niger-Congo & All & 1.9k \\
      Luganda   & lug & Niger-Congo & NER & 3.5k \\
      Luo   & luo &  & NER & 0.7k \\
      Bambara  & bam &  Mande & POS,Parsing & 4.4k \\
      Manx   & glv &  Indo-European & POS,Parsing & 37.6k \\
      Ancient Greek  & grc &  Indo-European & POS,Parsing & 8.0k \\
      Swiss German & gsw &  Indo-European & POS,Parsing & 2.5k \\
      Erzya & myv &  Uralic & POS,Parsing & 7.4k \\
    \bottomrule
    \end{tabular}}
    \caption{Languages used for evaluation.}
    \label{tab:languages}
\end{table}

\paragraph{Languages} For each task, we select languages that have task data but are not covered by the mBERT pretraining data. The languages we use can be found in \autoref{tab:languages}.
Most fall under the Few-Text setting~\citep{joshi-etal-2020-state}. We employ the same languages to simulate the No-Text setting as well.

\paragraph{Model} We use the multilingual BERT model~(mBERT) because it has competitive performance on under-represented languages~\citep{mad-x}. We find that our mBERT performance on WikiNER and POS is generally comparable or exceeds the XLM-R large results in \citet{adapt-1600-languages}. We additionally verify our results also hold for XLM-R in \autoref{sec:analysis}.

\subsection{Adaptation Data}
\paragraph{Lexicon} We extract lexicons between English and each target language from the PanLex database.\footnote{\url{https://panlex.org/snapshot/}} The number of lexicon entries varies from about 0.5k to 30k, and most of the lexicons have around 5k entries. The lexicon statistics for each language can be found in \autoref{tab:languages}.   

\paragraph{Pseudo Monolingual Data} English Wikipedia articles are used to synthesize monolingual data. We first tokenize the English articles using Stanza~\citep{qi2020stanza} and keep the first 200k sentences. To create pseudo monolingual data for a given target language, we replace each English word with its translation if the word exists in the bilingual lexicon. We randomly sample a target word if the English word has multiple possible translations because it is difficult to estimate translation probabilities due to lack of target text. 

\paragraph{Pseudo Labeled Data} Using the English training data for each task, we simply replace each English word in the labeled training data with its corresponding translation and retain its original label. For the sake of simplicity, we only use lexicon entries with a single word. 

\begin{table*}[]
    \centering
    \resizebox{\textwidth}{!}{
    \begin{tabular}{l|l|l|rr|rr|rr|rr|rr}
    \toprule
      & Method & Lexicon & WikiNER & $\Delta$ & MasakhaNER & $\Delta$ &  POS & $\Delta$ & Parsing & $\Delta$ & Avg. & $\Delta$ \\
     \midrule 
   \multirow{6}{*}{No-Text}  & mBERT & - & 47.6 & - & 46.1 & - & 36.1 & - & 16.5 & - & 36.5 & - \\
      \cmidrule{2-13}
       & Pseudo Trans-train & PanLex & 49.8 & \underline{2.2} & 54.4 & 8.3 & \textbf{51.1} & \underline{15.0} & 25.9 & \underline{9.4} & \textbf{45.2}$^*$ & \underline{8.7} \\
       & Pseudo MLM & PanLex & 49.8 & \underline{2.2} & 52.6 & 6.5 & 48.9 & 12.8 & 25.2 & 8.7 & 44.1$^*$ & 7.6 \\
       & Both & PanLex & 48.5 & 0.9 & \textbf{54.6} & \underline{8.5} & 48.7 & 12.6 & 25.9 & \underline{9.4} & 44.4$^*$ & 7.9 \\
       & Both+Label Distillation & PanLex & \textbf{50.6} & 2.1 & 53.5 & -1.1 & 50.3 & 1.6 & \textbf{26.0} & 0.1 & 45.1$^*$ & 0.7 \\
     \midrule 
     \midrule 
    \multirow{8}{*}{Few-Text} & Gold MLM & - & 49.5 & - & 53.6 & - & 60.6 & - & 40.2 & - & 50.9 & - \\
      \cmidrule{2-13}
       & Pseudo Trans-train & PanLex & 50.2 & 0.7 & 59.4 & \underline{5.8} & 59.3 & -1.3 & 37.0 & -3.2 & 51.4 & 0.5 \\
      \cmidrule{2-13}
       & Pseudo MLM & PanLex  & 50.7 & \underline{1.2} & 57.4 & 3.8 & 65.4 & \underline{4.8} & \textbf{43.5} & \underline{3.3} & 54.2$^*$ & \underline{3.3} \\
       &  & PanLex+Induced  & 52.2 & 1.5 & 58.5 & 0.9 & 64.7 & -0.7 & 41.5 & -2.0 & 54.2$^*$ & 0.0 \\
      \cmidrule{2-13}
       & Both & PanLex  & 50.1 & 0.6 & 59.2 & 5.6 & 60.7 & 0.1 & 38.3 & -1.9 & 52.0$^*$ & 1.1 \\
       &   & PanLex+Induced  & 52.6 & 2.5 & \textbf{61.1} & 1.9 & 59.5 & -1.2 & 35.3 & -3.0 & 52.0$^\dagger$ & 0.0 \\
      \cmidrule{2-13}
       & Both+Label Distillation & PanLex & 51.7 & 1.6 & 58.4 & -0.8 & \textbf{66.2} & 5.5 & 41.9 & 3.6 & 54.5$^*$ & 2.5 \\
       &  & PanLex+Induced & \textbf{53.2} & 1.5 & 59.4 & 1.0 & 65.8 & -0.4 & 40.7 & -1.2 & \textbf{54.7}$^*$ & 0.2 \\
    \bottomrule
    \end{tabular}}
    \caption{Average F1 score for languages in each task. We record F1 of the LAS for Parsing. We compare three adaptation methods~($\Delta$ indicates gains over baselines): Pseudo Trans-train, Pseudo MLM, and Both. We also examine two data refinement methods: Label Distillation~($\Delta$ is gains over Both) and PanLex+Induced~($\Delta$ is gains over PanLex). 
    \textbf{Bold} is the best result for each dataset, and \underline{underline} indicates the best improvements among the three adaptation methods over the baselines. We test the significance of the average gains over the baselines in the last column using paired bootstrap resampling. * indicates significant gains with $p<0.001$ and $\dagger$ indicates significant gains with $p<0.05$.} 
    \label{tab:main_results}
\end{table*}

\section{No-Text Setting\label{sec:g0_exp}}
We analyze the results of the following adaptation methods for the setting where we do not have any monolingual data. 

\paragraph{Pseudo MLM} The mBERT model is trained on the pseudo monolingual data using the MLM objective. We train the model for 5k steps for the NER tasks and 10k steps for the POS tagging and Parsing tasks. 


\paragraph{Pseudo Trans-train} We fine-tune mBERT or the model adapted with Pseudo MLM for a downstream task on the concatenation of both the English labeled data and the pseudo labeled data.

\paragraph{Label Distillation} We use the model adapted with Pseudo MLM as the teacher model to generate new labels for the pseudo labeled data, which we use jointly with the English labeled data to fine-tune the final model.

\subsection{Results}
The average performance of different adaptation methods averaged across all languages in each task can be found in \autoref{tab:main_results}.   
\paragraph{Pseudo Trans-train is the best method for No-Text.} Pseudo MLM and Pseudo Trans-train can both bring significant improvements over the mBERT baseline for all tasks. Pseudo Trans-train leads to the best aggregated result across all tasks, and it is also the best method or very close to the best method for each task. Adding Pseudo Trans-train on top of Pseudo MLM does not add much improvement. Label Distillation generally leads to better performance, but overall it is comparable to only using Pseudo Trans-train. 
\section{Few-Text Setting\label{sec:g1_exp}}
We test same adaptation methods introduced in \autoref{sec:g0_exp} for the Few-Text setting where we have a small amount of gold data. First we introduce the additional data and adaptation methods for this setting.

\subsection{Gold Data}
\paragraph{Gold Monolingual Data} We use the JHU Bible Corpus~\citep{jhu-bible-corpus} as the monolingual data. Following the setup in \citet{adapt-1600-languages}, we use the verses from the New Testament, which contain 5000 to 8000 sentences for each target language. 

\paragraph{Gold Parallel Data}
We can use the parallel data between English and the target languages from the Bible to extract additional word pairs. We use an existing unsupervised word alignment tool, eflomal~\citep{Ostling2016efmaral}, to generate word alignments for each sentence in the parallel Bible data. To create high quality lexicon entries, we only keep the word pairs that are aligned more than once, resulting in about 2k extra word pairs for each language. We then augment the PanLex lexicons with the induced lexicon entries.  

\subsection{Adaptation Methods}
\paragraph{Gold MLM} The mBERT model is trained on the gold monolingual Bible data in the target language using the MLM objective. Following the setting in \citet{adapt-1600-languages}, we train for 40 epochs for the NER task, and 80 epochs for the POS and Parsing tasks. 

\paragraph{Pseudo MLM} We conduct MLM training on both the Bible monolingual data and the pseudo monolingual data in the target language. The Bible data is up-sampled to match the size of the pseudo monolingual data. We train the model for 5k steps for the NER task and 10k steps for the POS tagging and Parsing tasks.


\subsection{Results}
The average performance in each task for Few-Text can be found in \autoref{tab:main_results}.
\paragraph{Pseudo MLM is the competitive strategy for Few-Text.} Unlike the No-Text setting, Pseudo Trans-train only marginally improves or even decreases the performance for three out of the four datasets we consider. On the other hand, Pseudo MLM, which uses both gold and pseudo monolingual data for MLM adaptation, consistently and significantly improves over Gold MLM for all tasks. Again, using Pseudo Trans-train on top of Pseudo MLM does not help and actually leads to relatively large performance loss for the syntactic tasks, such as POS tagging and Parsing.

\paragraph{Label Distillation brings significant improvements for the two syntactic tasks.} Notably, it is the best performing method for POS tagging, but it still lags behind Pseudo MLM for Parsing. This is likely because Parsing is a much harder task than POS tagging to generate correct labels. The effect of Label Distillation on the NER task is less consistent---it improves over Pseudo Trans-train for WikiNER but not for MasakhaNER. This is because the named entity tags of the same words in different languages likely remain the same so that the pseudo task data probably has less noise for Label Distillation to have consistent benefits.

\paragraph{Adding Induced Lexicons}
We examine the effect of using the lexicons augmented by word pairs induced from the Bible parallel data. The results can be found in \autoref{tab:main_results}. Adding the induced lexicon significantly improves the NER performance, while it hurts the two syntactic tasks. 

\begin{figure}
    \centering
    \includegraphics[width=0.48\columnwidth]{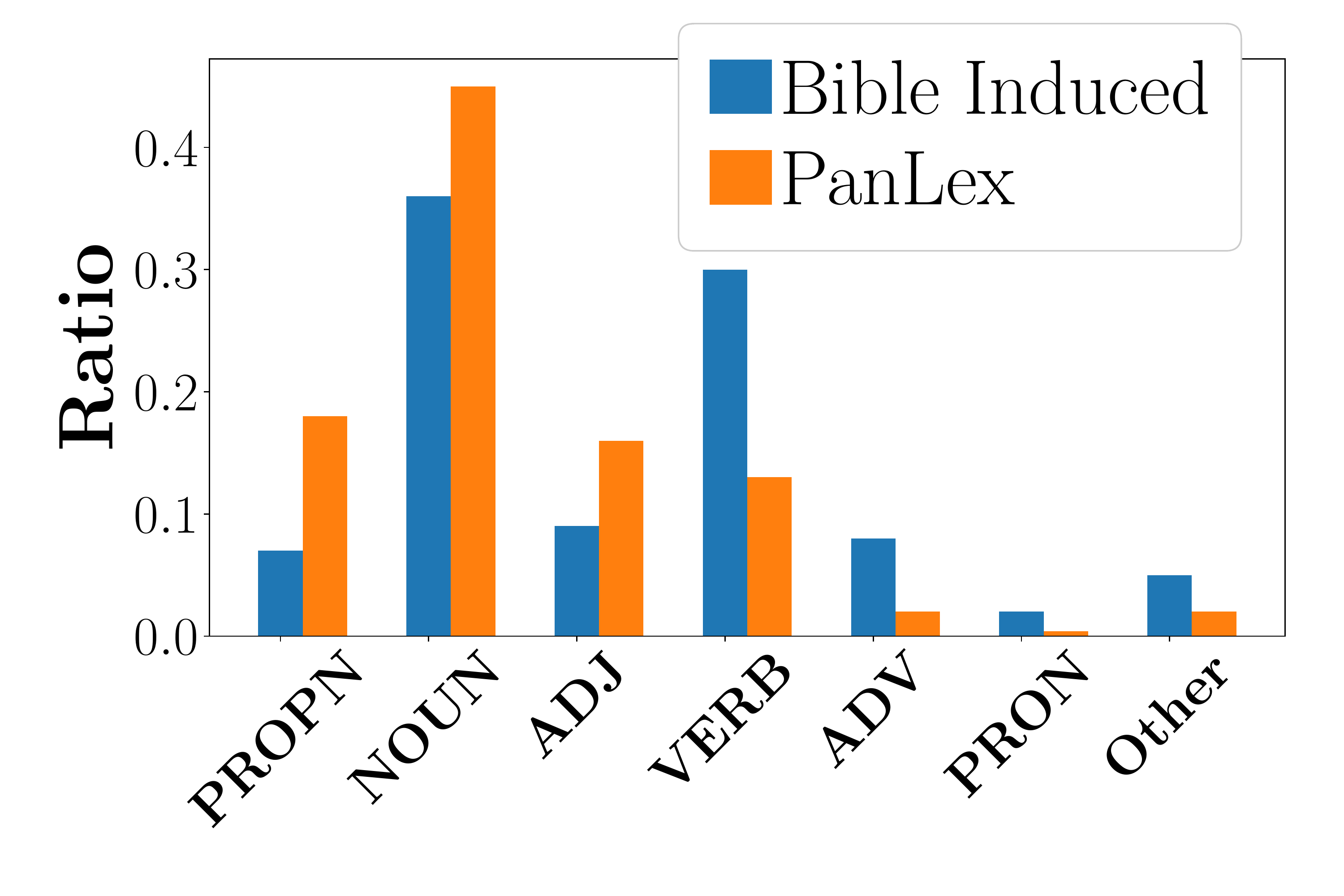}
    \includegraphics[width=0.48\columnwidth]{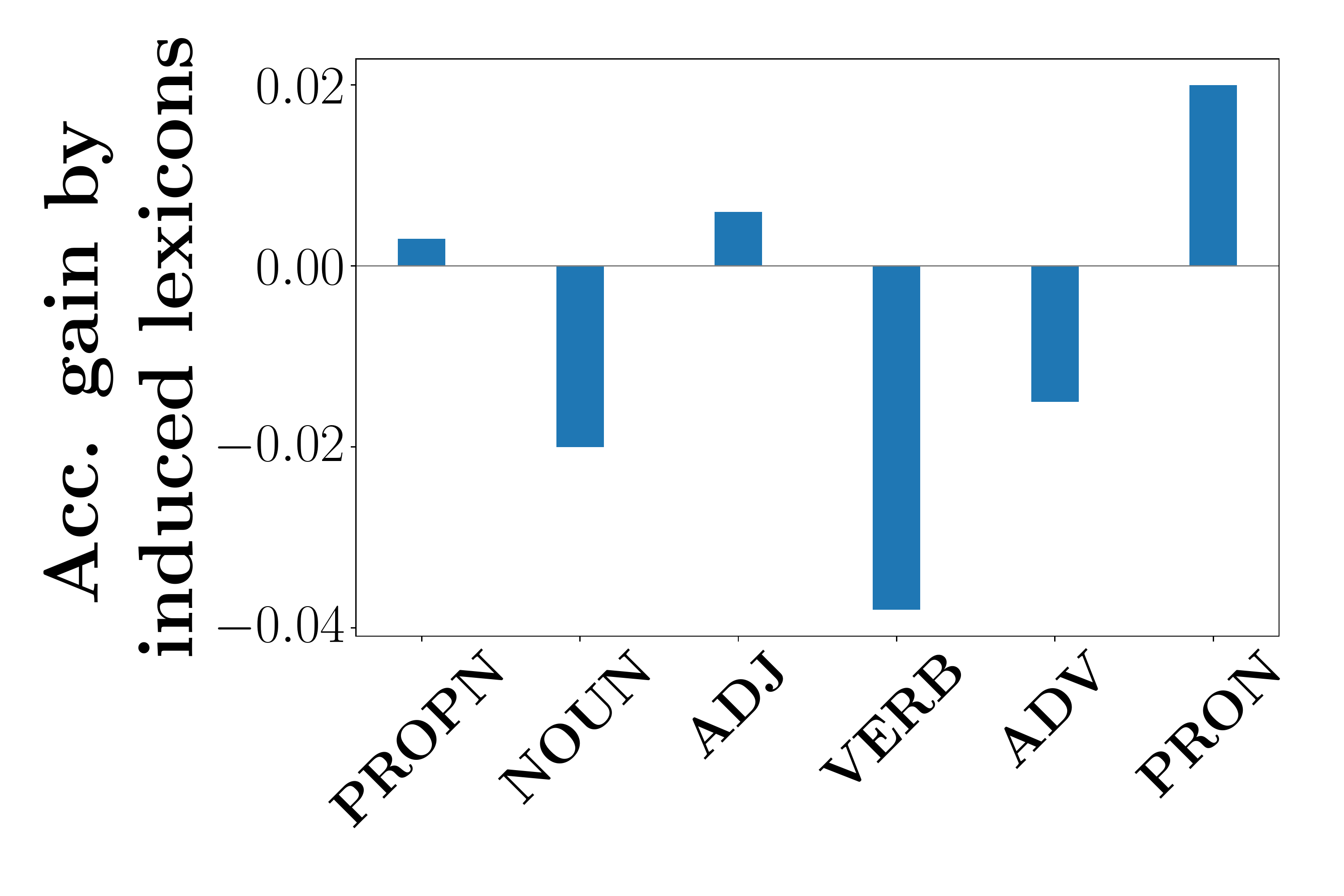}
    \caption{\textit{left}: Ratio of words with different POS tags in each lexicon. \textit{right}: POS accuracy gain of test words with different POS tags by using induced lexicons. The induced lexicons have more verbs but lead to worse performance on verbs. }
    \label{fig:lex_pos}
    \vspace{-1em}
\end{figure}

To understand what might have prevented the syntactic tasks from benefiting from the extra lexicon entries, we plot the distribution of the part-of-speech tags of the words in PanLex lexicons and the lexicons induced from the Bible in \autoref{fig:lex_pos}. PanLex lexicons have more nouns than the Bible lexicons while the Bible lexicons cover more verbs than PanLex. However, the higher verb coverage in induced lexicons actually leads to a larger prediction accuracy drop for verbs in the POS tagging task. We hypothesize that the pseudo monolingual data created using the induced lexicons would contain more target language verbs with the wrong word order, which could be more harmful for syntactic tasks than tasks that are less sensitive to word order such as NER.

\paragraph{Discrepancies between the two NER datasets} While WikiNER, along with POS tagging and Parsing, benefit the most from Pseudo MLM for Few-Text, MasakhaNER achieves the best result with Pseudo Trans-train. One possible explanation is that MasakhaNER contains data from the news domain, while WikiNER is created from Wikipedia. The pseudo monolingual data used for MLM is created from English Wikipedia articles, which could benefit WikiNER much more than MasakhaNER. On the other hand, the English NER training data for MasakhaNER is from the news domain, which potentially makes Pseudo Trans-train a stronger method for adapting the model simultaneously to the target language and to the news domain. One advantage of Pseudo MLM is that the English monolingual data is much cheaper to acquire, while Pseudo Trans-train is constrained by the amount of labeled data for a task. We show in \autoref{app:task_data_size} that Pseudo MLM has more benefit for MasakhaNER when we use a subset of the NER training data.

\section{Analyses\label{sec:analysis}}

\begin{table}[]
    \centering
    \resizebox{0.8\columnwidth}{!}{
    \begin{tabular}{l|c|c|c|c}
    \toprule
      & bam & glv & mlt & myv \\
     \midrule
      Gold MLM~(Ours) & 59.7 & 64.1 & 58.5 & 70.6 \\
      \citet{adapt-1600-languages}  & 60.5 & 59.7 & 59.6 & 66.6 \\
      \midrule 
      +Pseudo Trans-train & 57.4 & 63.2 & 69.1 & 63.8 \\
      +Pseudo MLM & 68.5 & 67.5 & \textbf{72.3} & 73.8 \\
      +Both & 60.3 & 64.5 & 69.3 & 65.9 \\
      +Both(Label Distillation) & \textbf{69.4} & \textbf{68.8} & 72.1 & \textbf{74.3} \\
    \bottomrule
    \end{tabular}}
    \caption{Results for POS tagging with XLM-R. Our methods follow similar trend as on mBERT and they lead to significant gains compared to prior work.}
    \label{tab:xlmr_pos}
\end{table}
\paragraph{Performance with XLM-R}
We mainly use mBERT because it has competitive performance for under-represented languages and it is more computationally efficient due to the smaller size. Here we verify our methods have the same trend when used on a different model XLM-R~\citep{xlmr}.
We focus on a subset of languages in the POS tagging task for the Few-Text setting and the results are in \autoref{tab:xlmr_pos}. We use the smaller XLM-R base for efficiency, and compare to the best result in prior work, which uses XLM-R large~\citep{adapt-1600-languages}. \autoref{tab:xlmr_pos} shows that our baseline is comparable or better than prior work. Similar to the conclusion in \autoref{sec:g1_exp}, Pseudo MLM is the competitive strategy that brings significant improvements over prior work. While adding Pseudo Trans-train to Pseudo MLM does not help, using Label Distillation further improves the performance.

\paragraph{Effect of Baseline Performance}
\begin{figure}
    \centering
    \includegraphics[width=0.45\columnwidth]{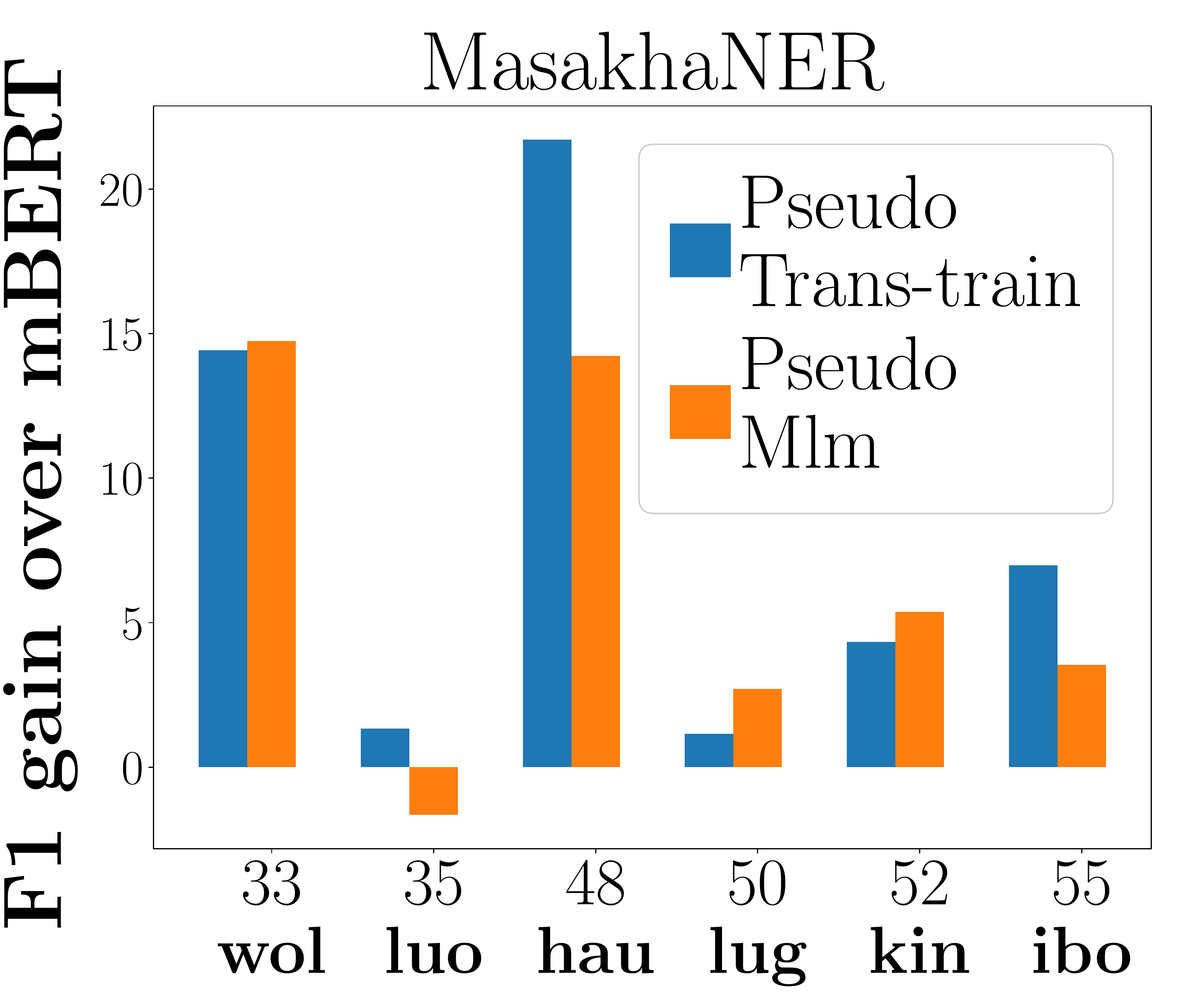}
    \includegraphics[width=0.45\columnwidth]{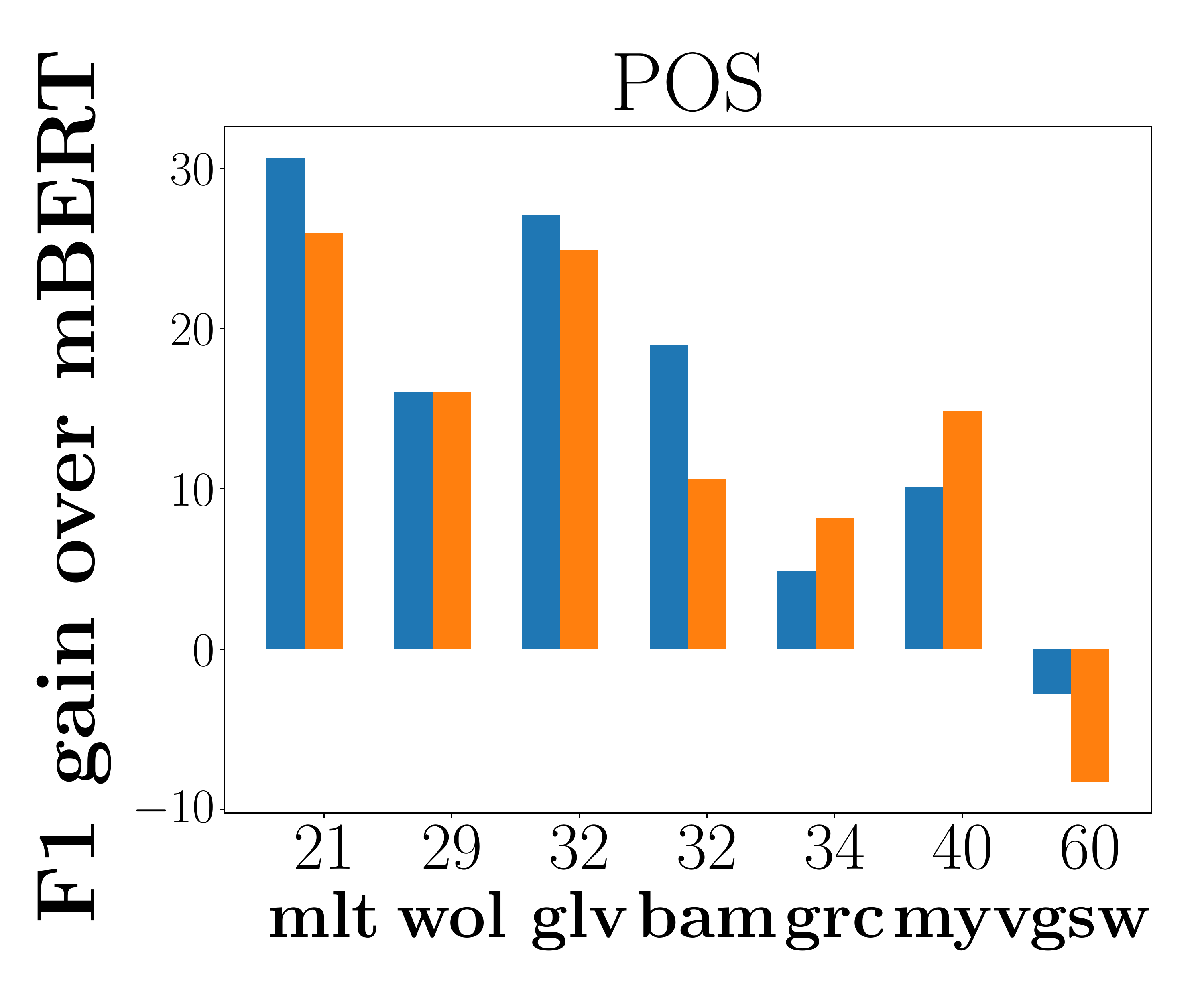}
    \caption{F1 gain over the baselines for languages with increasing baseline performance from left to right. Pseudo data tends to help more for languages with lower performance.}
    \label{fig:gain_vs_perf}
    \vspace{-5mm}
\end{figure}
Using pseudo data might be especially effective for languages with lower performance. We plot the improvement of different languages over the baseline in \autoref{fig:gain_vs_perf}, where languages are arranged with increasing baseline performance from left to right. We mainly plot Pseudo MLM and Pseudo Trans-train for simplicity. \autoref{fig:gain_vs_perf} shows that for both resource settings, lower performing languages on the left tend to have more performance improvement by using pseudo data.

\paragraph{Using NMT Model to Synthesize Data\label{sec:nmt}}
One problem with the pseudo data synthesized using word-to-word translation is that it cannot capture the correct word order or syntactic structure in the target language. If we have a good NMT system that translates English into the target language, we might be able to get more natural pseudo monolingual data by translating the English sentences to the target language. 

Since the target languages we consider are usually not supported by popular translation services, we train our own NMT system by fine-tuning an open sourced many-to-many NMT model on the Bible parallel data from English to the target language (details in \autoref{app:nmt}). Instead of creating pseudo monolingual data using the lexicon, we can simply use the fine-tuned NMT model to translate English monolingual data into the target language. 

The results of using NMT as opposed to lexicon for Pseudo MLM on all four tasks can be found in \autoref{tab:nmt}. Unfortunately, NMT is consistently worse than word-to-word translation using lexicons. 
We find that the translated monolingual data tend to have repeated words and phrases that are common in the Bible data, although the source sentence is from Wikipedia. This is because the NMT model overfits to the Bible data, and it fails to generate good translation for monolingual data from a different domain such as Wikipedia.

\begin{table}[]
    \centering
    \resizebox{0.8\columnwidth}{!}{
    \begin{tabular}{l|c|c|c|c}
    \toprule
         & WikiNER & MasakaNER & POS & Parsing \\
    \midrule
     Lexicon  & \textbf{45.0} & \textbf{56.0} & \textbf{63.7} & \textbf{40.7} \\
     NMT  & 42.2 & 55.8 & 58.9 & 37.7 \\
    \bottomrule
    \end{tabular}}
    \caption{F1 of using Pseudo MLM for Few-Text. Synthesizing data with NMT is consistently worse.}
    \label{tab:nmt}
    \vspace{-5mm}
\end{table}

\paragraph{Comparison to Few-shot Learning}
\citet{lauscher-etal-2020-zero} found that using as few as 10 labeled examples in the target language can significantly outperform the zero-shot transfer baseline for languages included in mBERT. 
We focus on the zero-shot setting in this paper because the languages we consider have very limited data and it could be expensive or unrealistic to annotate data in every task for thousands of languages. Nonetheless, we experiment with $k$-shot learning to examine its performance on low-resource languages in the MasakhaNER task. \autoref{tab:few_shot} shows that using 10 labeled examples brings improvements over the mBERT baseline for a subset of the languages, and it is mostly worse than our best adapted model without using any labeled data. When we have access to 100 examples, few-shot learning begins to reach or exceed our zero-shot model. In general, few-shot learning seems to require more data to consistently perform well for under-represented languages while our adaptation methods bring consistent gains without any labeled data. Combining the best adapted model with few-shot learning leads to mixed results. More research is needed to understand the annotation cost and benefit of few-shot learning for low-resource languages.  

\begin{table}[]
    \centering
    \resizebox{0.9\columnwidth}{!}{
    \begin{tabular}{l|c|c|c|c|c|c}
    \toprule
       Method  & hau & wol & lug & ibo & kin & luo\\
    \midrule
       mBERT  & 48.7 & 33.9 & 50.9 & 55.2 & 52.4 & 35.3 \\
       Best Adapted  & 74.4 & \textbf{60.3} & \textbf{61.6} & 63.6 & \textbf{63.8} & 42.6 \\
    \midrule
       10-shot  & 44.5 & 49.1 & 52.7 & 56.2 & 51.2 & 46.2 \\
       100-shot  & 64.0 & 56.9 & 58.3 & \textbf{65.5} & 55.7 & \textbf{51.6} \\
       Best Adapt+100-shot  & \textbf{76.1} & 57.3 & 61.3 & 63.2 & 62.6 & 49.4 \\
    \bottomrule
    \end{tabular}}
    \caption{Results on MasakhaNER for $k$-shot learning. We compare to the zero-shot mBERT baseline and our best adapted model.}
    \label{tab:few_shot}
    \vspace{-6mm}
\end{table}

\section{Related Work}
Several methods have been proposed to adapt pretrained language models to a target language. Most of them rely on MLM training using monolingual data in the target languages~\citep{wang-etal-2020-extending,chau-etal-2020-parsing,muller-etal-2021-unseen,mad-x,adapt-1600-languages}, competitive NMT systems trained on parallel data~\citep{xtreme,latent-translation-cross-lingual}, or some amount of labeled data in the target languages~\citep{lauscher-etal-2020-zero}. These methods cannot be easily extended to low-resource languages with no or limited amount of monolingual data, which account for more than $80\%$ of the World's languages~\citep{joshi-etal-2020-state}.

Bilingual lexicons have been commonly used for learning cross-lingual word embeddings \cite{mikolov2013exploiting,ruder2019survey}. Among these, some work uses lexicons to synthesize pseudo bilingual \cite{gouws-sogaard-2015-simple,duong2016learning} or pseudo multilingual corpora \cite{ammar2016massively}.
\citet{mayhew-etal-2017-cheap} propose to synthesize task data for NER using bilingual lexicons. More recently, \citet{khemchandani-etal-2021-exploiting} synthesize monolingual data in Indian languages for adapting pretrained language models via MLM. \citet{Hu2021amber} argue that using bilingual lexicons for alignment hurts performance compared to word-level alignment based on parallel corpora. Such parallel corpora, however, are not available for truly under-represented languages. \citet{Reid2021} employ a dictionary denoising objective where a word is replaced with its translation into a random language with a certain probability. This can be seen as text-to-text variant of our approach applied to multilingual pre-training. \citet{akyurek-andreas-2021-lexicon} propose to use lexicons to improve the compositionality of NLP models. None of the above works provide a systematic study of methods that utilize lexicons and limited data resources for adapting pretrained language models to languages with no or limited text.

\section{Conclusion and Discussion}
We propose a pipeline that leverages bilingual lexicons, an under-studied resource with much better language coverage than conventional data, to adapt pretrained multilingual models to under-represented languages. Through comprehensive studies, we find that using synthetic data can significantly boost the performance of these languages while the best method depends on the data availability. Our results show that we can make concrete progress towards including under-represented languages into the development of NLP systems by utilizing alternative data sources.

Our work also has some limitations. Since we focus on different methods of using lexicons, we restrict experiments to languages in Latin script and only use English as the source language for simplicity. Future work could explore the effect of using different source languages and combining transliteration~\citep{muller-etal-2021-unseen} or vocabulary extension \cite{unks2021} with lexicon-based data augmentation for languages in other scripts. We also did not test the data augmentation methods on higher-resourced languages as MLM fine-tuning and translate-train are already effective in that setting and our main goal is to support the languages with little textual data. 
Nonetheless, it would be interesting to examine whether our methods can deliver gains for high-resource languages, especially for test data in specialized domains. 

We point to the following future directions: First, phrases instead of single word entries could be used to create pseudo data. Second, additional lexicons beyond PanLex could be leveraged.\footnote{We provide a list of resources in Appendix \ref{sec:bilingual_lexicons}.} Third, more effort could be spent on digitizing both existing monolingual data such as books \cite{gref2016publishing} and lexicons into a format easily accessible by NLP practitioners. Although PanLex already covers over 5000 languages, some language varieties have only as little as 10 words in the database, while there exist many paper dictionaries that could be digitized through technologies such as OCR~\citep{rijhwani-etal-2020-ocr}.\footnote{\url{https://panlex.org/source-list/} contains a list of undigitized dictionaries.} 
Lexicon collection is also relatively fast, which could be a more cost effective strategy to significantly boost the performance of many languages without lexicons. Finally, the quality of synthetic data could be improved by incorporating morphology. However, we find that there is virtually no existing morphological analysis data or toolkits for the languages we consider. Future work could aim to improve the morphological analysis of these low-resource languages.
\section*{Acknowledgements}

This work was supported in part by the National Science Foundation under Grant Numbers 1761548 and 2040926. XW was supported in part by an Apple Graduate Fellowship. The authors would like to thank Aditi Chaudhary, Arya McCarthy, Shruti Rijhwani for discussions about the project, and Daan van Esch for the general feedback and pointing out additional linguistic resources.

\bibliography{anthology,custom}

\begin{thebibliography}{45}
\expandafter\ifx\csname natexlab\endcsname\relax\def\natexlab#1{#1}\fi

\bibitem[{Adelani et~al.(2021)Adelani, Abbott, Neubig, D'souza, Kreutzer,
  Lignos, Palen-Michel, Buzaaba, Rijhwani, Ruder, Mayhew, Azime, Muhammad,
  Emezue, Nakatumba-Nabende, Ogayo, Aremu, Gitau, Mbaye, Alabi, Yimam, Gwadabe,
  Ezeani, Niyongabo, Mukiibi, Otiende, Orife, David, Ngom, Adewumi, Rayson,
  Adeyemi, Muriuki, Anebi, Chukwuneke, Odu, Wairagala, Oyerinde, Siro, Bateesa,
  Oloyede, Wambui, Akinode, Nabagereka, Katusiime, Awokoya, MBOUP,
  Gebreyohannes, Tilaye, Nwaike, Wolde, Faye, Sibanda, Ahia, Dossou, Ogueji,
  DIOP, Diallo, Akinfaderin, Marengereke, and Osei}]{adelani2021masakhaner}
David~Ifeoluwa Adelani, Jade Abbott, Graham Neubig, Daniel D'souza, Julia
  Kreutzer, Constantine Lignos, Chester Palen-Michel, Happy Buzaaba, Shruti
  Rijhwani, Sebastian Ruder, Stephen Mayhew, Israel~Abebe Azime, Shamsuddeen
  Muhammad, Chris~Chinenye Emezue, Joyce Nakatumba-Nabende, Perez Ogayo,
  Anuoluwapo Aremu, Catherine Gitau, Derguene Mbaye, Jesujoba Alabi, Seid~Muhie
  Yimam, Tajuddeen Gwadabe, Ignatius Ezeani, Rubungo~Andre Niyongabo, Jonathan
  Mukiibi, Verrah Otiende, Iroro Orife, Davis David, Samba Ngom, Tosin Adewumi,
  Paul Rayson, Mofetoluwa Adeyemi, Gerald Muriuki, Emmanuel Anebi, Chiamaka
  Chukwuneke, Nkiruka Odu, Eric~Peter Wairagala, Samuel Oyerinde, Clemencia
  Siro, Tobius~Saul Bateesa, Temilola Oloyede, Yvonne Wambui, Victor Akinode,
  Deborah Nabagereka, Maurice Katusiime, Ayodele Awokoya, Mouhamadane MBOUP,
  Dibora Gebreyohannes, Henok Tilaye, Kelechi Nwaike, Degaga Wolde, Abdoulaye
  Faye, Blessing Sibanda, Orevaoghene Ahia, Bonaventure F.~P. Dossou, Kelechi
  Ogueji, Thierno~Ibrahima DIOP, Abdoulaye Diallo, Adewale Akinfaderin, Tendai
  Marengereke, and Salomey Osei. 2021.
\newblock Masakhaner: Named entity recognition for african languages.
\newblock In \emph{TACL}.

\bibitem[{Akyurek and Andreas(2021)}]{akyurek-andreas-2021-lexicon}
Ekin Akyurek and Jacob Andreas. 2021.
\newblock Lexicon learning for few shot sequence modeling.
\newblock In \emph{ACL}, Online. Association for Computational Linguistics.

\bibitem[{Ammar et~al.(2016)Ammar, Mulcaire, Tsvetkov, Lample, Dyer, and
  Smith}]{ammar2016massively}
Waleed Ammar, George Mulcaire, Yulia Tsvetkov, Guillaume Lample, Chris Dyer,
  and Noah~A Smith. 2016.
\newblock Massively multilingual word embeddings.
\newblock \emph{arXiv preprint arXiv:1602.01925}.

\bibitem[{Anastasopoulos and Neubig(2019)}]{anastasopoulos-neubig-2019-pushing}
Antonios Anastasopoulos and Graham Neubig. 2019.
\newblock Pushing the limits of low-resource morphological inflection.
\newblock In \emph{Proceedings of the 2019 Conference on Empirical Methods in
  Natural Language Processing and the 9th International Joint Conference on
  Natural Language Processing (EMNLP-IJCNLP)}, Hong Kong, China. Association
  for Computational Linguistics.

\bibitem[{Bird(2020)}]{bird-2020-decolonising}
Steven Bird. 2020.
\newblock \href {https://doi.org/10.18653/v1/2020.coling-main.313}
  {Decolonising speech and language technology}.
\newblock In \emph{Proceedings of the 28th International Conference on
  Computational Linguistics}, pages 3504--3519, Barcelona, Spain (Online).
  International Committee on Computational Linguistics.

\bibitem[{Blasi et~al.(2021)Blasi, Anastasopoulos, and
  Neubig}]{blasi2021systematic}
Dami{\'a}n Blasi, Antonios Anastasopoulos, and Graham Neubig. 2021.
\newblock Systematic inequalities in language technology performance across the
  world's languages.
\newblock \emph{arXiv preprint arXiv:2110.06733}.

\bibitem[{Boerger(2017)}]{rapidwordbrenda}
Brenda Boerger. 2017.
\newblock \href
  {https://scholarspace.manoa.hawaii.edu/bitstream/10125/41988/1/41988-b.pdf}
  {Rapid word collection, dictionary production, and community well-being}.

\bibitem[{Caswell et~al.(2020)Caswell, Breiner, van Esch, and
  Bapna}]{language-id-wild}
Isaac Caswell, Theresa Breiner, Daan van Esch, and Ankur Bapna. 2020.
\newblock Language {ID} in the wild: Unexpected challenges on the path to a
  thousand-language web text corpus.
\newblock In \emph{COLING}.

\bibitem[{Chau et~al.(2020)Chau, Lin, and Smith}]{chau-etal-2020-parsing}
Ethan~C. Chau, Lucy~H. Lin, and Noah~A. Smith. 2020.
\newblock Parsing with multilingual {BERT}, a small corpus, and a small
  treebank.
\newblock In \emph{Findings of EMNLP 2020}.

\bibitem[{Conneau et~al.(2020)Conneau, Khandelwal, Goyal, Chaudhary, Wenzek,
  Guzm{\'a}n, Grave, Ott, Zettlemoyer, and Stoyanov}]{xlmr}
Alexis Conneau, Kartikay Khandelwal, Naman Goyal, Vishrav Chaudhary, Guillaume
  Wenzek, Francisco Guzm{\'a}n, Edouard Grave, Myle Ott, Luke Zettlemoyer, and
  Veselin Stoyanov. 2020.
\newblock Unsupervised cross-lingual representation learning at scale.
\newblock In \emph{ACL}.

\bibitem[{Conneau and Lample(2019)}]{xlm}
Alexis Conneau and Guillaume Lample. 2019.
\newblock Crosslingual language model pretraining.
\newblock In \emph{NeurIPS}.

\bibitem[{Devlin et~al.(2019)Devlin, Chang, Lee, and Toutanova}]{bert}
Jacob Devlin, Ming-Wei Chang, Kenton Lee, and Kristina Toutanova. 2019.
\newblock {BERT}: Pre-training of deep bidirectional transformers for language
  understanding.
\newblock In \emph{NAACL}.

\bibitem[{Duong et~al.(2016)Duong, Kanayama, Ma, Bird, and
  Cohn}]{duong2016learning}
Long Duong, Hiroshi Kanayama, Tengfei Ma, Steven Bird, and Trevor Cohn. 2016.
\newblock Learning crosslingual word embeddings without bilingual corpora.
\newblock \emph{arXiv preprint arXiv:1606.09403}.

\bibitem[{Ebrahimi and Kann(2021)}]{adapt-1600-languages}
Abteen Ebrahimi and Katharina Kann. 2021.
\newblock \href {https://doi.org/10.18653/v1/2021.acl-long.351} {How to adapt
  your pretrained multilingual model to 1600 languages}.
\newblock In \emph{ACL}, Online. Association for Computational Linguistics.

\bibitem[{Gippert et~al.(2006)Gippert, Himmelmann, Mosel
  et~al.}]{gippert2006essentials}
Jost Gippert, Nikolaus Himmelmann, Ulrike Mosel, et~al. 2006.
\newblock \emph{Essentials of language documentation}.
\newblock Mouton de Gruyter Berl{\'\i}n.

\bibitem[{Gouws and S{\o}gaard(2015)}]{gouws-sogaard-2015-simple}
Stephan Gouws and Anders S{\o}gaard. 2015.
\newblock \href {https://doi.org/10.3115/v1/N15-1157} {Simple task-specific
  bilingual word embeddings}.
\newblock In \emph{Proceedings of the 2015 Conference of the North {A}merican
  Chapter of the Association for Computational Linguistics: Human Language
  Technologies}, pages 1386--1390, Denver, Colorado. Association for
  Computational Linguistics.

\bibitem[{Gref(2016)}]{gref2016publishing}
Emily~Kennedy Gref. 2016.
\newblock \emph{Publishing in North American Indigenous Languages}.
\newblock Ph.D. thesis, University of London.

\bibitem[{Gururangan et~al.(2020)Gururangan, Marasovi{\'c}, Swayamdipta, Lo,
  Beltagy, Downey, and Smith}]{dont-stop-pretrain-gururangan-etal-2020}
Suchin Gururangan, Ana Marasovi{\'c}, Swabha Swayamdipta, Kyle Lo, Iz~Beltagy,
  Doug Downey, and Noah~A. Smith. 2020.
\newblock Don{'}t stop pretraining: Adapt language models to domains and tasks.
\newblock In \emph{ACL}, Online.

\bibitem[{Howard and Ruder(2018)}]{Howard2018}
Jeremy Howard and Sebastian Ruder. 2018.
\newblock \href {http://arxiv.org/abs/1801.06146} {{Universal Language Model
  Fine-tuning for Text Classification}}.
\newblock In \emph{Proceedings of ACL 2018}.

\bibitem[{Hu et~al.(2021)Hu, Johnson, Firat, Siddhant, and
  Neubig}]{Hu2021amber}
Junjie Hu, Melvin Johnson, Orhan Firat, Aditya Siddhant, and Graham Neubig.
  2021.
\newblock \href {http://arxiv.org/abs/2010.07972} {{Explicit Alignment
  Objectives for Multilingual Bidirectional Encoders}}.
\newblock In \emph{Proceedings of NAACL 2021}.

\bibitem[{Hu et~al.(2020)Hu, Ruder, Siddhant, Neubig, Firat, and
  Johnson}]{xtreme}
Junjie Hu, Sebastian Ruder, Aditya Siddhant, Graham Neubig, Orhan Firat, and
  Melvin Johnson. 2020.
\newblock Xtreme: A massively multilingual multi-task benchmark for evaluating
  cross-lingual generalization.
\newblock In \emph{ICML}.

\bibitem[{Joshi et~al.(2020)Joshi, Santy, Budhiraja, Bali, and
  Choudhury}]{joshi-etal-2020-state}
Pratik Joshi, Sebastin Santy, Amar Budhiraja, Kalika Bali, and Monojit
  Choudhury. 2020.
\newblock The state and fate of linguistic diversity and inclusion in the {NLP}
  world.
\newblock In \emph{ACL}, Online. Association for Computational Linguistics.

\bibitem[{Khemchandani et~al.(2021)Khemchandani, Mehtani, Patil, Awasthi,
  Talukdar, and Sarawagi}]{khemchandani-etal-2021-exploiting}
Yash Khemchandani, Sarvesh Mehtani, Vaidehi Patil, Abhijeet Awasthi, Partha
  Talukdar, and Sunita Sarawagi. 2021.
\newblock Exploiting language relatedness for low web-resource language model
  adaptation: {A}n {I}ndic languages study.
\newblock In \emph{ACL}, Online. Association for Computational Linguistics.

\bibitem[{Kondratyuk and Straka(2019)}]{udify-kondratyuk-straka-2019-75}
Dan Kondratyuk and Milan Straka. 2019.
\newblock 75 languages, 1 model: Parsing universal dependencies universally.
\newblock In \emph{EMNLP}, Hong Kong, China.

\bibitem[{Lauscher et~al.(2020)Lauscher, Ravishankar, Vuli{\'c}, and
  Glava{\v{s}}}]{lauscher-etal-2020-zero}
Anne Lauscher, Vinit Ravishankar, Ivan Vuli{\'c}, and Goran Glava{\v{s}}. 2020.
\newblock From zero to hero: {O}n the limitations of zero-shot language
  transfer with multilingual {T}ransformers.
\newblock In \emph{EMNLP}, Online. Association for Computational Linguistics.

\bibitem[{Mayhew et~al.(2017)Mayhew, Tsai, and Roth}]{mayhew-etal-2017-cheap}
Stephen Mayhew, Chen-Tse Tsai, and Dan Roth. 2017.
\newblock Cheap translation for cross-lingual named entity recognition.
\newblock In \emph{EMNLP}, Copenhagen, Denmark. Association for Computational
  Linguistics.

\bibitem[{McCarthy et~al.(2020)McCarthy, Wicks, Lewis, Mueller, Wu, Adams,
  Nicolai, Post, and Yarowsky}]{jhu-bible-corpus}
Arya~D. McCarthy, Rachel Wicks, Dylan Lewis, Aaron Mueller, Winston Wu, Oliver
  Adams, Garrett Nicolai, Matt Post, and David Yarowsky. 2020.
\newblock \href {https://aclanthology.org/2020.lrec-1.352} {The {J}ohns
  {H}opkins {U}niversity {B}ible corpus: 1600+ tongues for typological
  exploration}.
\newblock In \emph{LREC}, pages 2884--2892, Marseille, France. European
  Language Resources Association.

\bibitem[{Mikolov et~al.(2013)Mikolov, Le, and
  Sutskever}]{mikolov2013exploiting}
Tomas Mikolov, Quoc~V Le, and Ilya Sutskever. 2013.
\newblock Exploiting similarities among languages for machine translation.
\newblock \emph{arXiv preprint arXiv:1309.4168}.

\bibitem[{Muller et~al.(2021)Muller, Anastasopoulos, Sagot, and
  Seddah}]{muller-etal-2021-unseen}
Benjamin Muller, Antonios Anastasopoulos, Beno{\^\i}t Sagot, and Djam{\'e}
  Seddah. 2021.
\newblock \href {https://aclanthology.org/2021.naacl-main.38} {When being
  unseen from m{BERT} is just the beginning: Handling new languages with
  multilingual language models}.
\newblock In \emph{NAACL}, Online.

\bibitem[{Nivre et~al.(2018)Nivre, Abrams, Agi{\'c}, Ahrenberg, Antonsen,
  Aranzabe, Arutie, Asahara, Ateyah, Attia et~al.}]{nivre2018universal}
Joakim Nivre, Mitchell Abrams, {\v{Z}}eljko Agi{\'c}, Lars Ahrenberg, Lene
  Antonsen, Maria~Jesus Aranzabe, Gashaw Arutie, Masayuki Asahara, Luma Ateyah,
  Mohammed Attia, et~al. 2018.
\newblock Universal dependencies 2.2.

\bibitem[{{\"O}stling and Tiedemann(2016)}]{Ostling2016efmaral}
Robert {\"O}stling and J{\"o}rg Tiedemann. 2016.
\newblock \href {http://ufal.mff.cuni.cz/pbml/106/art-ostling-tiedemann.pdf}
  {Efficient word alignment with {M}arkov {C}hain {M}onte {C}arlo}.
\newblock \emph{Prague Bulletin of Mathematical Linguistics}, 106:125--146.

\bibitem[{Ott et~al.(2019)Ott, Edunov, Baevski, Fan, Gross, Ng, Grangier, and
  Auli}]{ott2019fairseq}
Myle Ott, Sergey Edunov, Alexei Baevski, Angela Fan, Sam Gross, Nathan Ng,
  David Grangier, and Michael Auli. 2019.
\newblock fairseq: A fast, extensible toolkit for sequence modeling.
\newblock In \emph{Proceedings of NAACL-HLT 2019: Demonstrations}.

\bibitem[{Pan et~al.(2017)Pan, Zhang, May, Nothman, Knight, and
  Ji}]{pan-etal-2017-cross}
Xiaoman Pan, Boliang Zhang, Jonathan May, Joel Nothman, Kevin Knight, and Heng
  Ji. 2017.
\newblock \href {https://doi.org/10.18653/v1/P17-1178} {Cross-lingual name
  tagging and linking for 282 languages}.
\newblock In \emph{ACL}, pages 1946--1958, Vancouver, Canada. ACL.

\bibitem[{Pfeiffer et~al.(2020)Pfeiffer, Vuli{\'c}, Gurevych, and
  Ruder}]{mad-x}
Jonas Pfeiffer, Ivan Vuli{\'c}, Iryna Gurevych, and Sebastian Ruder. 2020.
\newblock {MAD-X}: {A}n {A}dapter-{B}ased {F}ramework for {M}ulti-{T}ask
  {C}ross-{L}ingual {T}ransfer.
\newblock In \emph{EMNLP}, Online. Association for Computational Linguistics.

\bibitem[{Pfeiffer et~al.(2021)Pfeiffer, Vuli{\'{c}}, Gurevych, and
  Ruder}]{unks2021}
Jonas Pfeiffer, Ivan Vuli{\'{c}}, Iryna Gurevych, and Sebastian Ruder. 2021.
\newblock \href {http://arxiv.org/abs/2012.15562} {{UNKs Everywhere: Adapting
  Multilingual Language Models to New Scripts}}.
\newblock In \emph{Proceedings of EMNLP 2021}.

\bibitem[{Pires et~al.(2019)Pires, Schlinger, and
  Garrette}]{pires-etal-2019-multilingual}
Telmo Pires, Eva Schlinger, and Dan Garrette. 2019.
\newblock How multilingual is multilingual {BERT}?
\newblock In \emph{ACL}, Florence, Italy.

\bibitem[{Ponti et~al.(2021)Ponti, Kreutzer, Vulić, and
  Reddy}]{latent-translation-cross-lingual}
Edoardo~Maria Ponti, Julia Kreutzer, Ivan Vulić, and Siva Reddy. 2021.
\newblock Modelling latent translations for cross-lingual transfer.
\newblock In \emph{Arxiv}.

\bibitem[{Qi et~al.(2020)Qi, Zhang, Zhang, Bolton, and Manning}]{qi2020stanza}
Peng Qi, Yuhao Zhang, Yuhui Zhang, Jason Bolton, and Christopher~D. Manning.
  2020.
\newblock Stanza: A {Python} natural language processing toolkit for many human
  languages.
\newblock In \emph{ACL}.

\bibitem[{Rahimi et~al.(2019)Rahimi, Li, and Cohn}]{rahimi-etal-2019-massively}
Afshin Rahimi, Yuan Li, and Trevor Cohn. 2019.
\newblock \href {https://www.aclweb.org/anthology/P19-1015} {Massively
  multilingual transfer for {NER}}.
\newblock In \emph{ACL}, pages 151--164, Florence, Italy. Association for
  Computational Linguistics.

\bibitem[{Reid and Artetxe(2021)}]{Reid2021}
Machel Reid and Mikel Artetxe. 2021.
\newblock \href {http://arxiv.org/abs/2108.01887} {{PARADISE: Exploiting
  Parallel Data for Multilingual Sequence-to-Sequence Pretraining}}.
\newblock \emph{arXiv preprint arXiv:2108.01887}.

\bibitem[{Rijhwani et~al.(2020)Rijhwani, Anastasopoulos, and
  Neubig}]{rijhwani-etal-2020-ocr}
Shruti Rijhwani, Antonios Anastasopoulos, and Graham Neubig. 2020.
\newblock {OCR} {P}ost {C}orrection for {E}ndangered {L}anguage {T}exts.
\newblock In \emph{EMNLP}, Online. Association for Computational Linguistics.

\bibitem[{Ruder et~al.(2021)Ruder, Constant, Botha, Siddhant, Firat, Fu, Liu,
  Hu, Neubig, and Johnson}]{xtreme-r}
Sebastian Ruder, Noah Constant, Jan Botha, Aditya Siddhant, Orhan Firat, Jinlan
  Fu, Pengfei Liu, Junjie Hu, Graham Neubig, and Melvin Johnson. 2021.
\newblock \href {http://arxiv.org/abs/2104.07412} {{XTREME-R: Towards More
  Challenging and Nuanced Multilingual Evaluation}}.
\newblock In \emph{Proceedings of EMNLP 2021}.

\bibitem[{Ruder et~al.(2019)Ruder, Vuli{\'c}, and S{\o}gaard}]{ruder2019survey}
Sebastian Ruder, Ivan Vuli{\'c}, and Anders S{\o}gaard. 2019.
\newblock A survey of cross-lingual word embedding models.
\newblock \emph{Journal of Artificial Intelligence Research}, 65:569--631.

\bibitem[{Wang et~al.(2020)Wang, K, Mayhew, and
  Roth}]{wang-etal-2020-extending}
Zihan Wang, Karthikeyan K, Stephen Mayhew, and Dan Roth. 2020.
\newblock \href {https://aclanthology.org/2020.findings-emnlp.240} {Extending
  multilingual {BERT} to low-resource languages}.
\newblock In \emph{EMNLP-Findings}, Online. Association for Computational
  Linguistics.

\bibitem[{Wu and Dredze(2019)}]{wu-dredze-2019-beto}
Shijie Wu and Mark Dredze. 2019.
\newblock \href {https://www.aclweb.org/anthology/D19-1077} {Beto, bentz,
  becas: The surprising cross-lingual effectiveness of {BERT}}.
\newblock In \emph{EMNLP}.

\end{thebibliography}
\bibliographystyle{acl_natbib}

\clearpage
\appendix
\section{Appendix}
\subsection{Experiment Details\label{app:exp_detail}}
 For all experiments using MLM training for NER tasks, we train 5000 steps, or about equivalent to 40 epochs on Bible; for MLM training for POS tagging and Parsing, we train 10000 steps, or equivalent to 80 epochs on Bible. We use learning rate of $2e-5$, batch size of $32$, and maximum sequence length of $128$. We did not tune these hyperparameters because we mostly follow the ones provided in \citep{adapt-1600-languages}.

To finetune the model for a downstream task, we use learning rate of $2e-5$ and batch size of $32$. We train all models for 10 epochs and pick the checkpoint with the best performance on the English development set.

We use a single GPU for all adaptation and fine-tuning experiments. Pseudo MLM usually takes less than 5 hours. Pseudo Trans-train and other task specific fine-tuning usually takes around 2 to 3 hours.

\subsection{NMT Models\label{app:nmt}}
We use the many-to-many NMT models provided in the fairseq repoo~\citep{ott2019fairseq}. We use the model with $175$M parameters and finetune the NMT model for $50$ epochs on the parallel data from the Bible.

We use beam size of $5$ to generate translations.

\subsection{Induced lexicons help languages with Fewer PanLex Entries} 
\begin{figure}
    \centering
    \includegraphics[width=0.45\columnwidth]{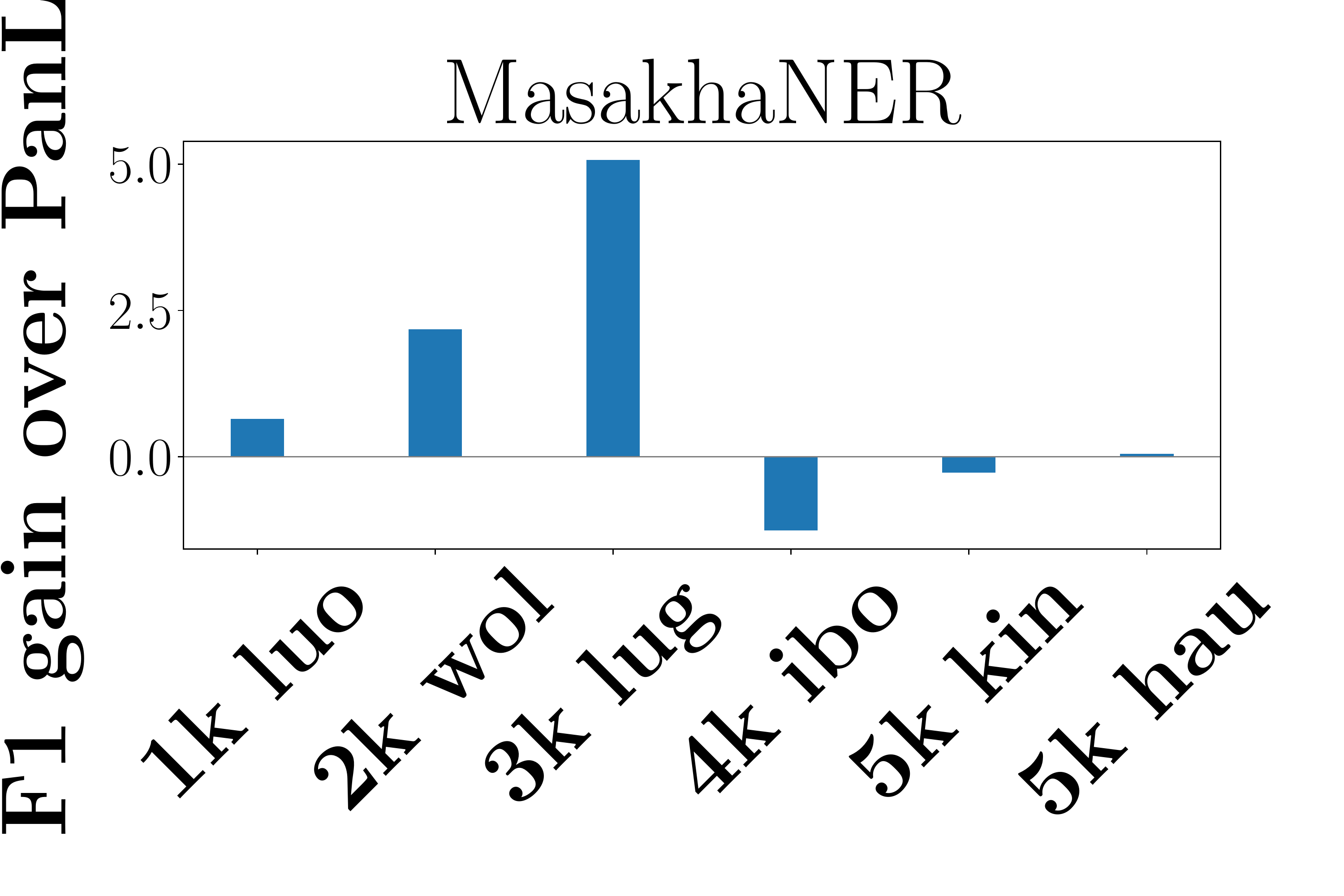}
    \includegraphics[width=0.45\columnwidth]{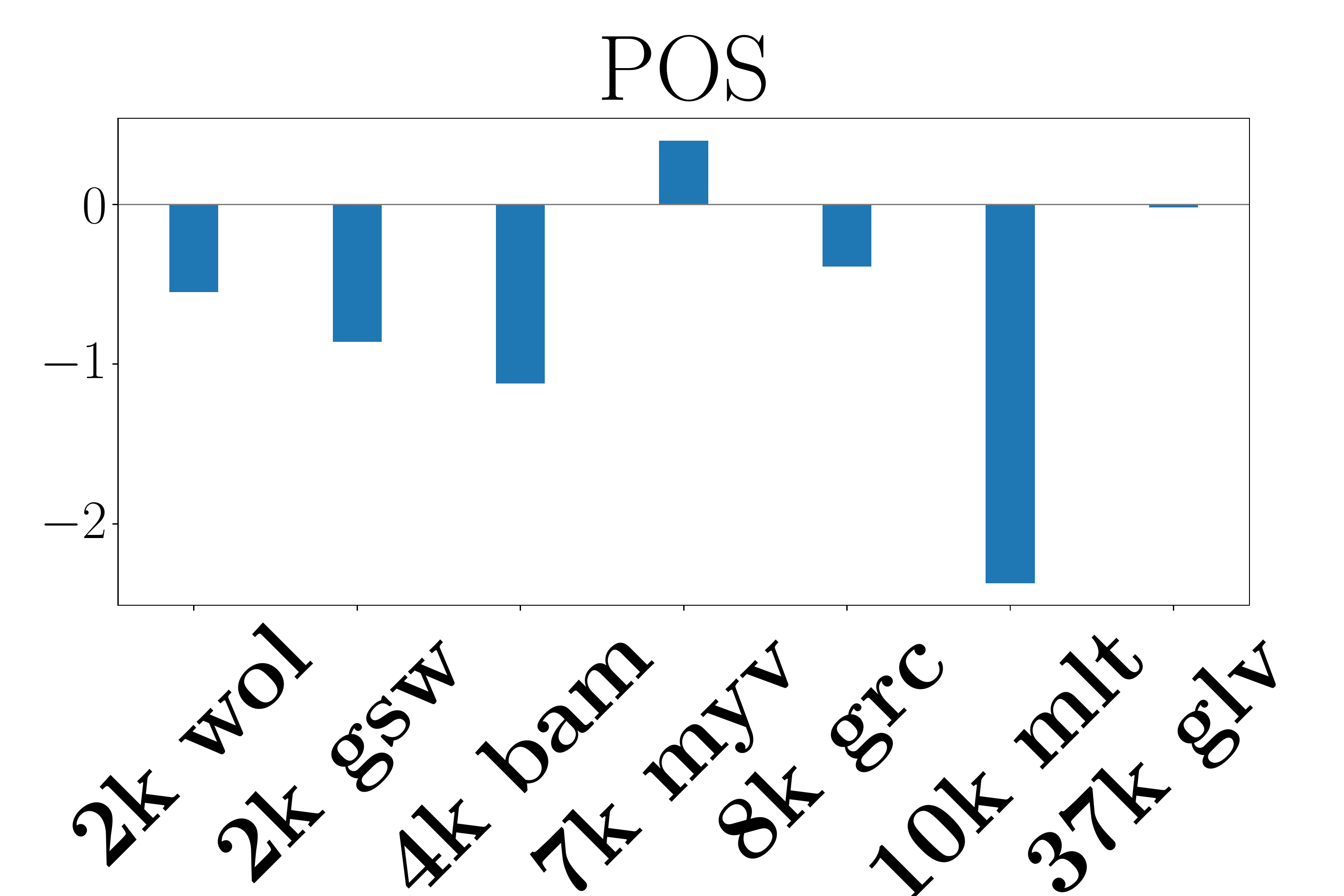}
    \caption{Improvements of using combined lexicons compared to PanLex lexicons for Pseudo MLM. Languages with fewer PanLex lexicons tend to benefit more from the combined lexicons.}
    \label{fig:gain_vs_lexc}
    \vspace{-5mm}
\end{figure}

We plot the performance difference between using combined lexicons and PanLex for the Few-Text in \autoref{fig:gain_vs_lexc}. The languages are arranged from left to right based on increasing amount of PanLex entries. For MasakhaNER, the three languages with fewer entries in PanLex have much more significant gains by using the combined lexicon. While using the combined lexicons generally hurts POS tagging, the languages with fewer entries in PanLex tend to have less performance decrease.

\subsection{Effect of Task Data Size\label{app:task_data_size}}
\begin{figure}
    \centering
    \includegraphics[width=0.4\columnwidth]{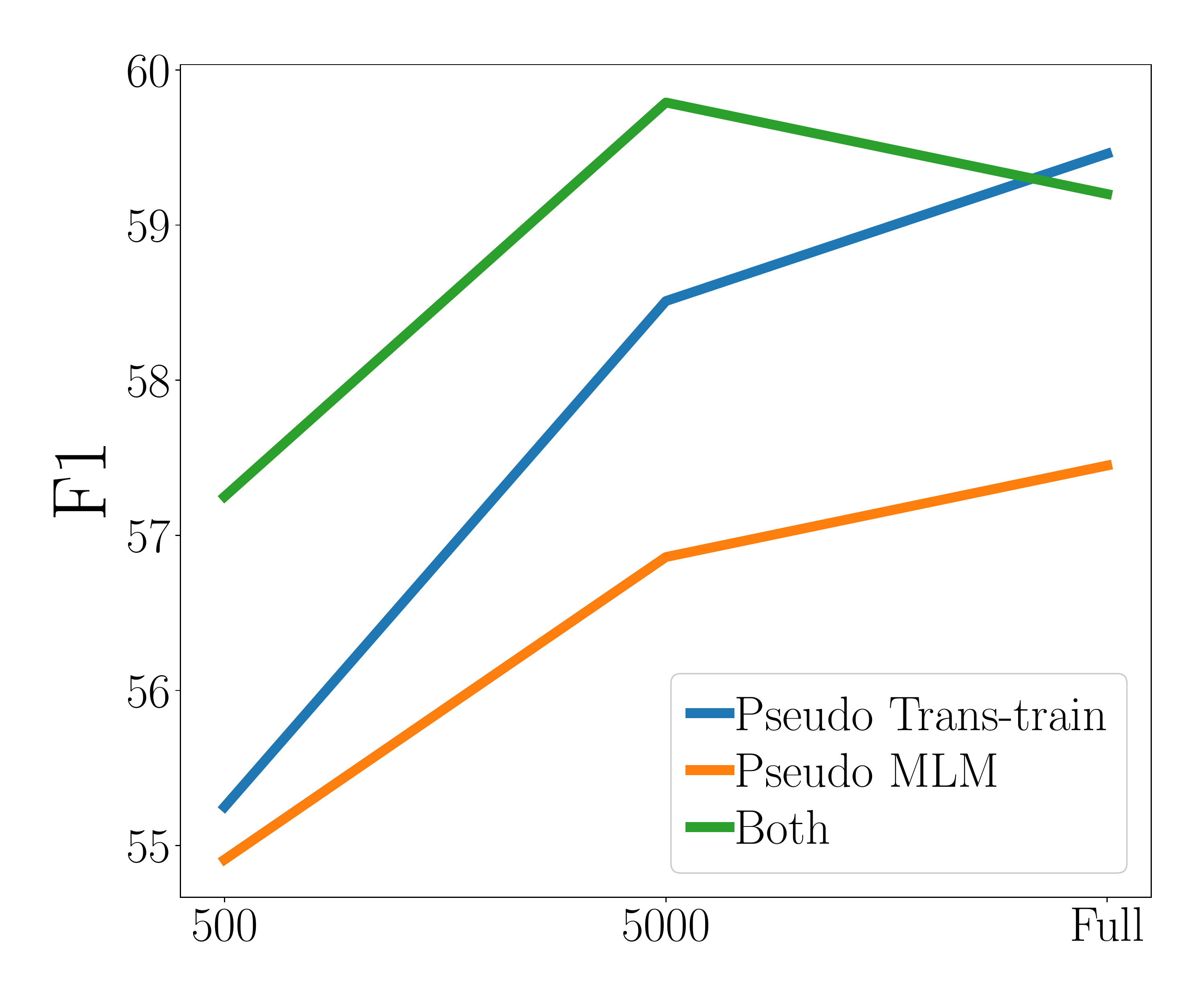}
    \caption{F1 on MasakhaNER with different amount of labeled data. Pseudo MLM becomes beneficial when the labeled training data is small.}
    \label{fig:fewshot}
\end{figure}
Our experiments in \autoref{tab:main_results} show that MasakhaNER benefits more from Pseudo Trans-train, likely because the labeled data is closer to the domain of the test data. 
However, this result might not hold when the amount of labeled data is limited. One advantage of Pseudo MLM over Pseudo Trans-train is that it only requires English monolingual data to synthesize pseudo training data, while Pseudo Trans-train is constrained by the availability of labeled data. We subsample the amount of English NER training data for MasakhaNER and plot the average F1 score of Pseudo Trans-train, pseudo MLM and using both. \autoref{fig:fewshot} shows that the advantage of Pseudo Trans-train on MasakhaNER decreases as the number of labeled data decreases, and using both methods is more competitive when the task data is small.   

\subsection{List of Bilingual Lexicons} \label{sec:bilingual_lexicons}

We provide a list of bilingual lexicons beyond PanLex:
\begin{itemize}
    \item Swadesh lists in about 200 languages in Wikipedia\footnote{\url{https://en.wiktionary.org/wiki/Appendix:Swadesh_lists}}
    \item Words in 3156 language varietities in CLICS\footnote{\url{https://clics.clld.org/}}
    \item Intercontinental Dictionary Series in about 300 languages\footnote{\url{https://ids.clld.org/}}
    \item 40-item wordlists in 5,000+ languages in ASJP\footnote{\url{https://asjp.clld.org/}}
    \item Austronesian Basic Vocabulary Database in 1,700+ languages\footnote{\url{https://abvd.shh.mpg.de/austronesian/}}
    \item Diachronic Atlas of Comparative Linguistics in 500 languages\footnote{\url{https://diacl.ht.lu.se/}}
\end{itemize}

\subsection{Lexicon Extraction}
We use a simple python script to extract the lexicons from the PanLex database, and directly use them for synthesizing the pseudo data. We will open-source the script in our codebase.

\subsection{Performance for Individual Language}
\newcommand{\rpm}{\raisebox{.2ex}{$\scriptstyle\pm$}}
\begin{table*}[]
    \centering
    \resizebox{\textwidth}{!}{
    \begin{tabular}{l|l|l|c|c|c|c|c|c}
    \toprule
      & Method & Lexicon & hau & wol  & lug & ibo & kin & luo \\
     \midrule 
   \multirow{6}{*}{No-Text}  & mBERT &   & 48.7 & 33.9 & 50.9  & 55.2 & 52.4 & 35.3 \\
      \cmidrule{2-9}
       & Pseudo Trans-train & PanLex & 70.4 & 48.3 & 52.1 & 62.2 &  56.8 & 36.6 \\
       & Pseudo MLM & PanLex &  62.9 & 48.7 & 53.6 & 58.7 & 57.8 & 33.7 \\
       & Both & PanLex & 69.5 & 52.6 & 55.3 & 62.3 & 57.3 & 31.9 \\
       & Both+Label Distillation & PanLex & 64.1 & 47.4 & 55.0 & 62.1 & 58.3 & 34.3 \\
     \midrule 
     \midrule 
    \multirow{8}{*}{Few-Text} & Gold MLM & & 54.3 & 48.4 & 59.8 & 58.4 & 58.2 & 42.7 \\
      \cmidrule{2-9}
       & Pseudo Trans-train & PanLex & 71.5 & 58.1 & 60.8 & 63.4 & 61.2 & 41.4  \\
      \cmidrule{2-9}
       & Pseudo MLM & PanLex  & 64.3 & 55.0 & 58.5 & 63.6 & 62.1 & 40.9  \\
       &  & PanLex+Induced  & 64.3 & 57.2 & 63.6 & 62.4 & 61.9 & 41.6 \\
      \cmidrule{2-9}
       & Both & PanLex  & 73.5 & 58.3 & 60.6 & 63.1 & 62.5 & 37.0 \\
       &   & PanLex+Induced  & 74.4 & 60.3 & 61.6 & 63.6 & 63.8 & 42.6 \\
      \cmidrule{2-9}
       & Both+Label Distillation & PanLex &  65.02 & 56.4 & 60.8 & 64.7 & 62.5 & 40.8 \\
       &  & PanLex+Induced & 66.8 & 56.1 & 62.7 & 63.6 & 64.2 & 43.2 \\
    \bottomrule
    \end{tabular}}
    \caption{Average F1 score for languages in MasakaNER} 
    \label{tab:ner_masa}
    \vspace{-1em}
\end{table*}

\begin{table*}[]
    \centering
    \resizebox{\textwidth}{!}{
    \begin{tabular}{l|l|l|r|r|r|r|r|r|r}
    \toprule
      & Method & Lexicon & bam & glv  & grc & gsw & mlt & myv & wol \\
     \midrule 
   \multirow{6}{*}{No-Text}  & mBERT &  & 32.8 & 32.5 & 34.9 & 60.8 & 21.8 & 40.4 & 29.2 \\
      \cmidrule{2-10}
       & Pseudo Trans-train & PanLex & 51.8 & 59.6 & 39.9 & 58.0 &  52.4 & 50.6 & 45.3 \\
       & Pseudo MLM & PanLex & 43.5 & 57.5 & 43.1 & 52.6 & 47.7 & 55.3 & 42.7 \\
       & Both & PanLex & 51.8 & 59.0 & 36.4 & 50.3 & 50.6 & 50.2 & 42.8 \\
       & Both+Label Distillation & PanLex & 45.3 & 59.0 & 43.0 & 54.3 & 49.8 & 56.1 & 44.4  \\
     \midrule 
     \midrule 
    \multirow{8}{*}{Few-Text} & Gold MLM & & 57.2 & 61.7 & 40.8 & 65.0 & 64.0 & 69.2 & 66.3 \\
      \cmidrule{2-10}
       & Pseudo Trans-train & PanLex & 56.8 & 62.2 & 44.9 & 62.8 & 61.6 & 63.4 & 63.1 \\
      \cmidrule{2-10}
       & Pseudo MLM & PanLex  &  66.5 & 64.3 & 48.6 & 67.1 & 70.4 & 72.1 & 68.9 \\
       &  & PanLex+Induced  & 65.4 & 64.3 & 48.2 & 66.3 & 68.1 & 72.5 & 68.4 \\
      \cmidrule{2-10}
       & Both & PanLex  &  59.5 & 63.0 & 42.1 & 65.2 & 63.1 & 65.9 & 66.4 \\
       &   & PanLex+Induced  & 60.4 & 63.0 & 42.2 & 62.8 & 60.1 & 70.3 & 57.6 \\
      \cmidrule{2-10}
       & Both+Label Distillation & PanLex & 66.9 & 65.3 & 50.1 & 68.5 & 71.0 & 72.5 & 69.5 \\
       &  & PanLex+Induced & 65.6 & 64.7 & 49.7 & 68.9 & 70.0 & 72.9 & 69.3 \\
    \bottomrule
    \end{tabular}}
    \caption{Average F1 score for languages in UDPOS} 
    \label{tab:ner_masa}
    \vspace{-1em}
\end{table*}

\begin{table*}[]
    \centering
    \resizebox{\textwidth}{!}{
    \begin{tabular}{l|l|l|r|r|r|r|r|r|r}
    \toprule
      & Method & Lexicon & bam & glv  & grc & gsw & mlt & myv & wol \\
     \midrule 
   \multirow{6}{*}{No-Text}  & mBERT & & 10.5 & 8.4 & 17.4 & 45.2 & 7.7 & 16.9 & 9.7 \\
      \cmidrule{2-10}
       & Pseudo Trans-train & PanLex & 15.2 & 41.1 & 19.3 & 31.7 & 35.0 & 22.4 & 16.6 \\
       & Pseudo MLM & PanLex & 15.4 & 39.5 & 20.6 & 30.7 & 28.2 & 25.9 & 16.5 \\
       & Both & PanLex & 16.3 & 42.0 & 17.4 & 30.1 & 33.3 & 24.5 & 17.5 \\
       & Both+Label Distillation & PanLex & 16.5 & 41.6 & 20.1 & 29.8 & 31.8 & 26.1 & 16.2 \\
     \midrule 
     \midrule 
    \multirow{8}{*}{Few-Text} & Gold MLM & & 25.1 & 43.2 & 21.9 & 49.8 & 50.4 & 44.9 & 46.0 \\
      \cmidrule{2-9}
       & Pseudo Trans-train & PanLex & 22.4 & 43.9 & 24.4 & 42.4 & 48.3 & 38.9 & 39.0 \\
      \cmidrule{2-9}
       & Pseudo MLM & PanLex  & 31.2 & 50.0 & 25.9 & 50.5 & 53.1 & 45.9 & 48.1 \\
       &  & PanLex+Induced  & 28.9 & 48.9 & 23.9 & 44.3 & 50.5 & 46.7 & 47.5 \\
      \cmidrule{2-9}
       & Both & PanLex  &  23.2 & 45.3 & 20.7 & 45.5 & 49.9 & 39.5 & 44.2 \\
       &   & PanLex+Induced  & 24.5 & 45.2 & 20.3 & 37.7 & 48.2 & 38.8 & 32.0 \\
      \cmidrule{2-9}
       & Both+Label Distillation & PanLex & 29.1 & 50.4 & 24.6 & 46.8 & 52.1 & 44.3 & 45.8  \\
       &  & PanLex+Induced & 28.2 & 50.4 & 24.4 & 40.7 & 51.6 & 45.7 & 43.7 \\
    \bottomrule
    \end{tabular}}
    \caption{Average F1 score for languages in Parsing} 
    \label{tab:ner_masa}
    \vspace{-1em}
\end{table*}

\begin{table*}[]
    \centering
    \resizebox{\textwidth}{!}{
    \begin{tabular}{l|l|l|r|r|r|r|r|r|r|r|r|r}
    \toprule
      & Method & Lexicon & ace & bak  & crh & hak & ibo & ilo & kin & mhr & mlt & mri \\
     \midrule 
   \multirow{6}{*}{No-Text}  & mBERT &  & 39.4  & 57.9 & 48.2 & 28.5 & 41.7 & 59.8 & 57.3 & 47.7 & 53.1 & 42.7 \\
      \cmidrule{2-13}
       & Pseudo Trans-train & PanLex & 41.1 & 63.2 & 47.1 & 30.9 & 49.4 & 62.8 & 56.7 & 49.9 & 63.4 & 32.7  \\
       & Pseudo MLM & PanLex & 38.4 & 60.1 & 46.9 & 30.2 & 46.8 & 62.4 & 60.2 & 51.8 & 59.3 & 42.5 \\
       & Both & PanLex & 38.8 & 57.2 & 43.9 & 30.2 & 48.5 & 63.3 & 57.4 & 51.1 & 62.8 & 32.1 \\
       & Both+Label Distillation & PanLex & 38.4 & 59.3 & 46.4 & 32.3 & 48.6 & 65.8 & 62.7 & 51.5 & 64.1 & 36.6  \\
     \midrule 
     \midrule 
    \multirow{8}{*}{Few-Text} & Gold MLM & & 38.7 & 57.9 & 48.4 & 37.2 & 48.0 & 60.5 & 56.4 & 51.4 & 64.5 & 32.2 \\
      \cmidrule{2-13} 
       & Pseudo Trans-train & PanLex & 38.2 & 60.9 & 48.6 & 37.0 & 50.1 & 63.6 & 56.9 & 52.7 & 62.4 & 32.0  \\
      \cmidrule{2-13}
       & Pseudo MLM & PanLex  & 41.4 & 58.2 & 47.7 & 36.0 & 50.7 & 65.7 & 61.4 & 50.5 & 62.4 & 33.3 \\
       &  & PanLex+Induced  & 43.3 & 57.5 & 47.8 & 37.6 & 47.4 & 66.9 & 59.6 & 53.1 & 63.5 & 45.2  \\
      \cmidrule{2-13}
       & Both & PanLex  & 41.5 & 57.9 & 47.8 & 35.5 & 50.0 & 65.4 & 56.9 & 50.9 & 62.4 & 32.4 \\
       &   & PanLex+Induced  & 40.7 & 57.6 & 51.3 & 40.2 & 48.8 & 67.4 & 60.5 & 56.8 & 65.3 & 37.3 \\
      \cmidrule{2-13}
       & Both+Label Distillation & PanLex & 46.0 & 56.5 & 50.0 & 35.3 & 49.6 & 65.1 & 61.5 & 52.6 & 65.9 & 34.8  \\
       &  & PanLex+Induced & 45.7 & 61.6 & 52.1 & 38.7 & 49.1 & 63.6 & 63.0 & 55.2 & 66.9 & 36.3 \\
    \bottomrule
    \end{tabular}}
    \caption{Average F1 score for languages in WikiNER} 
    \label{tab:ner_masa}
    \vspace{-1em}
\end{table*}

\end{document}